\documentclass[sigconf, nonacm]{acmart}

\renewcommand\footnotetextcopyrightpermission[1]{} %

\makeatletter
\def\@oddfoot{\hfil \small \textit{Accepted to PACMMOD / SIGMOD 2027. This is the author's accepted manuscript.} \hfil}
\def\@evenfoot{\hfil \small \textit{Accepted to PACMMOD / SIGMOD 2027. This is the author's accepted manuscript.} \hfil}
\makeatother

\usepackage{multicol}     
\usepackage{appendix}     
\usepackage{acronym}
\usepackage{xcolor}

\usepackage{eso-pic}
\usepackage{microtype}
\usepackage{makecell}
\usepackage{booktabs}
\usepackage{graphicx}
\usepackage{flushend}
\usepackage{caption}
\usepackage{tabularx}
\usepackage{subcaption}
\usepackage{amsfonts}
\usepackage{amsmath}
\usepackage{amsthm}
\usepackage{algorithm}
\usepackage{algorithmic}
\usepackage{multirow}
\usepackage{color}
\usepackage{xcolor}
\usepackage{booktabs}
\usepackage{float}
\usepackage{balance}
\usepackage{xspace}
\usepackage{threeparttable}
\usepackage{url}
\usepackage{enumitem}
\usepackage{siunitx}

\usepackage{amssymb}
\usepackage[utf8]{inputenc}
\usepackage{booktabs} 
\usepackage{longtable} 
\usepackage{amsthm}
\usepackage{marginnote}
\newtheorem{Def}{Definition}
\newtheorem*{Pro*}{Problem} 
\AtBeginDocument{%
  }
\acrodef{AutoML}{Automatic Machine Learning}
\acrodef{RL}{Reinforcement Learning}
\acrodef{TD}{Temporal Difference}
\acrodef{TB}{Trajectory Balance}
\acrodef{FiLM}{Feature-wise Linear Modulation}
\acrodef{GFlowNets}{Generative Flow Networks}
\acrodef{C-GFlowNets}{Conditional Generative Flow Networks}
\acrodef{SOTA}{state-of-the-art}
\acrodef{LLM}{Large Language Model}

\author{Kunyu Ni}
\affiliation{%
  \institution{Ocean University of China}
  \country{China}
}
\email{kyni@stu.ouc.edu.cn}

\author{Lei Cao}
\affiliation{%
  \institution{University of Arizona}
  \country{USA}
}
\email{lcao@csail.mit.edu}

\author{Jie He}
\affiliation{%
  \institution{University of Science and Technology Beijing}
  \country{China}
}
\email{hejie@ustb.edu.cn}

\author{Xiaotong Zhang}
\affiliation{%
  \institution{University of Science and Technology Beijing}
  \country{China}
}
\email{zxt@ustb.edu.cn}

\author{Jianfeng Jin}
\affiliation{%
  \institution{Northeastern University}
  \country{China}
}
\email{jinjf@atm.neu.edu.cn}

\author{Junyu Dong}
\affiliation{%
  \institution{Ocean University of China}
  \country{China}
}
\email{dongjunyu@ouc.edu.cn}

\author{Yanwei Yu}
\authornote{Corresponding author.}
\affiliation{%
  \institution{Ocean University of China}
  \country{China}
}
\email{yuyanwei@ouc.edu.cn}

\definecolor{darkred}{RGB}{139,0,0}

\usepackage{xcolor}
\usepackage{multicol}
\usepackage{appendix}
\usepackage[normalem]{ulem}

\colorlet{R1blue}{black}
\colorlet{R2orange}{black}
\colorlet{R3purple}{black}
\colorlet{R4teal}{black}
\colorlet{MetaRed}{black}

\newcommand{\reva}[1]{{\color{R1blue}#1}}   
\newcommand{\revb}[1]{{\color{R2orange}#1}}  
\newcommand{\revc}[1]{{\color{R3purple}#1}} 
\newcommand{\revm}[1]{{\color{MetaRed}#1}}  

\newcommand{\ie}{\textit{i}.\textit{e}.}
\newcommand{\eg}{\textit{e.g.},\xspace}

\newlist{compactitem}{itemize}{3} 
\setlist[compactitem]{label=\textbullet, nosep, leftmargin=0cm,itemindent=.5cm}
\renewcommand{\shortauthors}{Ni et al.}

\begin{document}
\AddToShipoutPictureFG*{%
  \AtPageUpperLeft{%
    \raisebox{-18pt}[0pt][0pt]{%
      \makebox[\paperwidth][c]{\large \itshape
      Accepted to SIGMOD 2027. This is the author's accepted manuscript.}%
    }%
  }%
}
\title{FlowPipe: LLM-Enhanced Conditional Generative Flow Networks for Data Preparation Pipeline Construction
}

\renewcommand{\shortauthors}{Ni et al.}

\begin{abstract}
Data preparation pipelines are a primary mechanism for improving data quality in machine learning workflows, transforming raw, error-prone tables into learning-ready data via sequential cleaning and feature transformation operators. However, automated pipeline construction remains computationally prohibitive due to the combinatorial complexity of operator sequences and the high cost of end-to-end evaluation. While Reinforcement Learning provides a principled discrete search paradigm, \ac{SOTA} Multi-DQN architectures suffer from three fundamental limitations: \textit{structural dissonance}, where decoupled value estimators hinder long-horizon credit assignment; \textit{semantic detachment}, where dataset context is treated as a superficial additive bias rather than strictly conditioning the agent's reasoning; and \textit{exploration inefficiency} in a vast, sparse optimization landscape with many invalid pipeline states. To address these challenges, we propose \textbf{FlowPipe}, a unified framework that reformulates pipeline synthesis as conditional probabilistic flow generation over a directed acyclic graph. First, FlowPipe employs a \ac{C-GFlowNets} optimized via a Trajectory Balance objective, establishing a direct gradient path from terminal validation rewards to early actions to ensure holistic credit assignment. Second, to resolve semantic detachment, we introduce \textit{Deep Semantic Modulation} via \ac{FiLM}, which allows LLM-derived logical priors to multiplicatively modulate the policy's internal activation maps, thereby structurally adapting the decision logic to the dataset context. Finally, to mitigate exploration inefficiency, we incorporate failure awareness into the flow objective to prune semantically invalid states early and concentrate search mass on high-potential regions. Extensive experiments on two benchmark suites comprising 74 real-world datasets show that FlowPipe significantly outperforms \ac{SOTA} baselines, improving accuracy by an average of \textbf{11.96\%} while achieving a \textbf{12.5$\times$} speedup in training convergence. The source code is available at \url{https://github.com/KunyuNi/FlowPipe}.
    
\end{abstract}

\maketitle

\vspace{-2mm}
\section{Introduction}

Data quality has emerged as the decisive factor in the lifecycle of machine learning (ML) systems. In real applications, the performance of these systems is largely bounded by the veracity and integrity of the underlying data~\cite{grinsztajn2022tree,shwartz2022tabular}. Turning raw, noisy datasets into high quality, ML ready data requires rigorous \textit{data preparation pipelines} that orchestrate heterogeneous operators to cleanse artifacts, impute missing values, and rectify distributional anomalies~\cite{abedjan2016detecting,diffprep}. 

Designing an effective data preparation pipeline is challenging due to the vast combinatorial space of available cleaning and transformation operators. So far it remains a labor-intensive process that relies on human expertise~\cite{zaidi2017market,sculley2015hidden,polyzotis2019data}. For example, domain experts utilize rich \textit{semantic context}—such as attribute definitions, domain constraints, and valid value ranges—to orchestrate operator selection, ensuring that cleaning decisions align with the intrinsic logic of the data~\cite{haipipe}. 

If there were a system that could automate this process, it would be to the great need of numerous ML-driven applications. However, this is challenging. The optimization landscape is combinatorial and non-convex, and naively applying operators without semantic awareness often fails to resolve underlying quality issues or, worse, introduces new artifacts (e.g., invalid imputations)~\cite{autoweka,TPOT}. The complexity is further exacerbated by the high computational burden of evaluation, which depends on delayed supervision signals from downstream models rather than explicit ground truth. Consequently, practitioners frequently default to generic configurations or inefficient trial-and-error strategies~\cite{feurer2015efficient,li2018hyperband}, leaving critical data quality problems unresolved.

\begin{figure*}[t]
    \centering
    \includegraphics[width=1\textwidth,height=6cm]{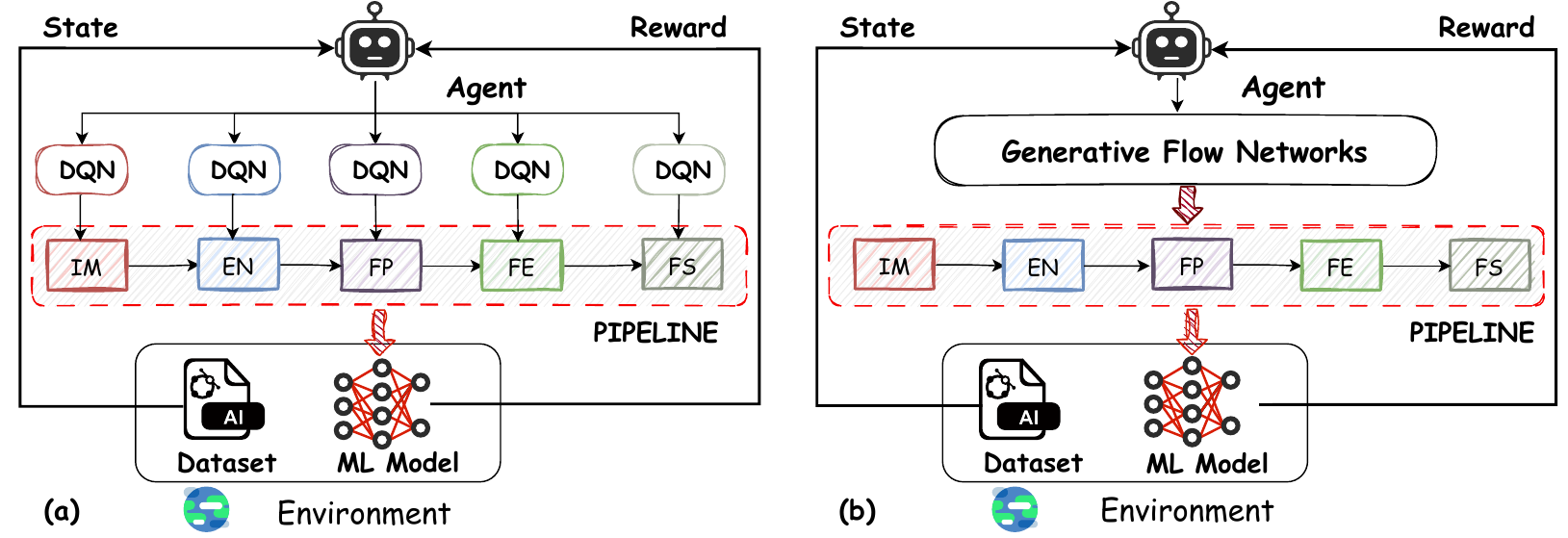}    
    \vspace{-6mm}
\caption{
\textbf{Paradigm Comparison.} 
(a) \textbf{The Multi-DQN architecture} employs decoupled agents for each stage, causing structural dissonance and credit assignment problems. 
(b) \textbf{FlowPipe} models the entire pipeline synthesis as a coherent probabilistic flow via GFlowNets, ensuring holistic credit assignment. 
The target data preparation pipeline consists of five sequential stages: Imputer (IM), Encoder (EN), Feature Preprocessing (FP), Feature Engineering (FE), and Feature Selection (FS).
}
\label{fig:paradigm_comparison}
\vspace{-4mm}
\end{figure*}

Recent work has extended \ac{AutoML} beyond hyperparameter tuning and architecture search to \emph{automated data preparation pipeline construction}~\cite{feurer2015efficient,learn2clean,yang2019oboe,schelter2018automating}.
Existing systems broadly follow two paradigms: \emph{differentiable pipelines} and \emph{discrete construction}. Differentiable approaches~\cite{liudarts,li2020differentiable,fan2020autofs} relax operator selection into continuous parameters to enable gradient-based optimization. However, this method is inherently constrained by the non-differentiable nature of many essential cleaning operators and incurs prohibitive computational overhead due to repetitive end-to-end gradient updates. On the other hand, discrete search methods, such as evolutionary algorithms TPOT~\cite{TPOT} and Bayesian optimization Auto-Weka~\cite{autoweka}, treat pipeline evaluation as a monolithic black-box problem. These approaches often struggle to capture the fine-grained, state-dependent transitions required to assemble complex, multi-step workflows.

Consequently, \ac{RL} has emerged as the premier framework within the discrete paradigm. By formulating pipeline construction as a sequential decision process, \ac{RL} naturally handles the discrete operators that differentiable methods cannot. Moreover, it explicitly captures the step-wise dependencies that monolithic evolutionary search algorithms often overlook. This approach perfectly aligns with data preparation tasks, where the utility of an operator is dictated by the context created by preceding transformations.

However, translating this theoretical alignment into a robust system remains a significant challenge. Pioneer RL-based works, such as DeepLine~\cite{deepline}, modeled pipeline generation using a unified DQN agent. While this monolithic design simplifies the architecture, it struggles within the high-dimensional action space, where the combinatorial complexity of operators and parameters grows exponentially. To address this, state-of-the-art (SOTA) systems like HaiPipe~\cite{haipipe} and CtxPipe~\cite{ctxpipe} adopt a Multi-DQN architecture, as shown in Figure~\ref{fig:paradigm_comparison}(a). This approach isolates the decision logic for each component of the pipeline, such as Imputation, Encoding, and Feature Engineering, into separate networks. Each DQN is tasked with the decision-making for a specific pipeline component, facilitating more efficient management of the vast search space.

\textbf{Challenges.} Despite these advances, existing solutions face three fundamental obstacles that limit their effectiveness in real-world scenarios:

\begin{compactitem}[leftmargin=*]
\item \textbf{Credit Assignment Problem.} 
The core challenge lies in distinguishing the specific contribution of individual upstream actions to the final downstream performance.
While the Multi-DQN formulation effectively decomposes the combinatorial action space into manageable sub-spaces, this design creates a fundamental \textit{structural dissonance}. Although the operational sequence remains sequentially interdependent, the decision-making is distributed across isolated Q-networks. This separation makes it difficult to consistently attribute delayed validation rewards to upstream pipeline decisions.

\item \textbf{Semantic Detachment.}
Beyond the structural separation, the architecture is further limited by \textit{Semantic Detachment}, where the dataset's semantic context fails to actively inform the agent's core reasoning process. While CtxPipe~\cite{ctxpipe} pioneers the inclusion of dataset context, it mechanically injects this information as an \textit{extrinsic bias term} added to the final value estimates. This integration strategy causes the semantic signals to bypass the agent's core reasoning machinery. Operationally, the agent evaluates the pipeline state using a fixed, generic logic while the context serves merely as an independent value adjustment applied at the output layer. Consequently, the policy remains structurally unconditioned by the dataset profile and fails to adapt its fundamental cleaning strategy to the specific semantic context of the target dataset.

\item \textbf{Exploration Inefficiency.}
This refers to the prohibitive computational cost of navigating an optimization landscape where the search space grows exponentially with the sequence length.
Current approaches exacerbate this challenge by relying solely on shallow features like table schema or statistical metadata to guide exploration. Lacking richer semantic cues to prune this vast combinatorial manifold, and given that validation requires expensive end-to-end training, the agent is forced into a brute-force trial-and-error loop. Consequently, standard strategies suffer from severe sample inefficiency, often failing to locate optimal regions within a reasonable time budget.
\end{compactitem}

\textbf{Proposed Approach.} To bridge the gap between RL's theoretical promise and its practical implementation, we propose FlowPipe, a system that formulates pipeline synthesis as a unified conditional probabilistic flow generation process over a linear compositional space. First, to resolve the \textit{Credit Assignment Problem}, FlowPipe departs from the fragmented Multi-DQN architecture by deploying a \ac{C-GFlowNets} optimized via the Trajectory Balance objective. Unlike short-sighted step-wise updates that isolate decisions, our formulation explicitly models the probability flow of the complete pipeline trajectory. This establishes a direct gradient path from the terminal validation reward back to earliest actions, ensuring holistic supervision across long horizons. Simultaneously, to resolve {\it Semantic Detachment}, we discard the extrinsic bias approach of prior works that restricts context integration to a superficial concatenation at the final output. 
Instead, we employ {\it Deep Semantic Modulation} via \ac{FiLM} layers to condition the agent’s intermediate representations on LLM-enhanced semantic context. By functioning as a dynamic filter, this mechanism selectively amplifies relevant feature signals while dampening incompatible ones deep within the network layers. Consequently, the model implicitly re-configures its core reasoning logic, ensuring that the decision-making process is structurally adapted to the unique constraints of each target dataset. 
Finally, to mitigate \textit{Exploration Inefficiency}, we implement a {\it Failure-aware Trajectory Balance} objective. We incorporate failure awareness into the flow objective to prune invalid pipeline states early. This effectively anchors the exploration budget, pruning the exponential combinatorial space to concentrate search mass exclusively on semantically valid, high-potential regions. 
Extensive experiments on 74 real-world datasets confirm FlowPipe's dominance over \ac{SOTA} baselines, delivering an 11.96\% average accuracy gain with 12.5$\times$ faster training. Crucially, FlowPipe matches the quality of a 30-hour-per-dataset exhaustive search in mere seconds.


In summary, our key contributions are as follows:
\begin{itemize}[leftmargin=*]
    \item \textbf{Generative Flow Formulation.} We propose \textbf{FlowPipe}, the first framework to formulate data preparation as a conditional generative flow. By modeling the complete pipeline as a cohesive unit, our Trajectory Balance objective establishes a direct gradient path for precise credit attribution across long-horizon dependencies.
    
    \item \textbf{Semantic-Constrained Policy.} We introduce a mechanism that conditions pipeline generation on LLM-derived logical priors. By modulating the policy network via \ac{FiLM}, we enforce strict semantic consistency between data properties and operator logic.

    \item \textbf{Efficiency and Effectiveness.} We conduct extensive experiments across the DiffPrep and DeepLine benchmarks. FlowPipe not only establishes a new \ac{SOTA} by improving accuracy by \textbf{11.17\%} and \textbf{12.17\%}, respectively, but also demonstrates superior computational efficiency. Compared to the best-performed baseline (\ie, CtxPipe), FlowPipe achieves inference speedups of \textbf{20.99\%} on DiffPrep and \textbf{26.55\%} on DeepLine.  
    
\end{itemize}

\vspace{-2.5mm}
\section{Related Work}
\subsection{Automated Pipeline Construction}
\label{sec:related_dp}
Automated data preparation pipeline construction is a fundamental problem in data management. Unlike standard hyperparameter optimization, it involves searching a combinatorial space of heterogeneous operators subject to logical dependencies~\cite{chu2016data,lee2017big}. Prior work mainly follows two paradigms: \emph{discrete construction}, which synthesizes pipelines via sequential decisions over an operator graph, and \emph{differentiable construction}, which relaxes discrete choices into continuous parameters for gradient-based optimization. {\it FlowPipe adopts the discrete paradigm and navigates the resulting search space using generative flow modeling.}

\textbf{Discrete Construction.} 
Discrete methods frame pipeline synthesis as sequential decision-making over a finite operator set~\cite{shang2019democratizing}. Early systems such as Auto-Weka~\cite{autoweka} and Auto-Sklearn~\cite{feurer2015efficient} couple Bayesian optimization with meta-learning to jointly optimize preprocessing and modeling, but often reduce data preparation to a shallow, mostly fixed stage. TPOT~\cite{TPOT} uses genetic programming to evolve tree-structured pipelines, yet can struggle to converge in high-dimensional spaces due to limited directed guidance.  
\revb{Similarly, SAGA~\cite{saga} advances this category by employing a specialized heuristic evolutionary search for pipeline generation.} 
More generally, evolutionary and Bayesian approaches tend to evaluate pipelines monolithically, making it difficult to model fine-grained, state-dependent transitions typical of multi-step cleaning.

To better capture sequential dependencies, \ac{RL}-based methods have become prominent. Learn2Clean~\cite{learn2clean} and DeepLine~\cite{deepline} model long-horizon decision processes but often rely on rigid heuristic constraints (\eg fixed operator ordering) for tractability. For more complex settings, HAIPipe~\cite{haipipe} and CtxPipe~\cite{ctxpipe} adopt Multi-DQN designs that decompose pipeline generation into stages handled by separate agents; however, this decoupling can hinder global credit assignment because each Q-network optimizes a local objective. In CtxPipe, despite incorporating dataset context, reliance on random snapshots and shallow additive fusion may limit the ability to capture deep semantic dependencies between data and operators. More recently, SwiftDP~\cite{swiftdp} combines MCTS with meta-learning, but its reliance on meta-features from seen datasets limits generalization. ShapleyPipe~\cite{chang2025shapleypipe} uses Shapley-based search, yet incurs thousands of pipeline evaluations and offers limited transfer. Neither explicitly exploits dataset semantics for pruning and adaptation.

\textbf{Differentiable Construction.} Conversely, this paradigm relaxes the discrete search problem into continuous optimization. WindTunnel~\cite{yu2021windtunnel} and DiffML~\cite{hilprecht2023diffml} model pipelines as differentiable tensor computations or weighted mixtures, allowing gradients to backpropagate from the downstream model. DiffPrep\cite{diffprep} further formalizes this by enabling the learning of transformation hyperparameters via differentiable relaxations. Although these methods theoretically allow for fine-grained optimization, they are  constrained to differentiable operators and incur substantial computational overhead due to the requirement of repetitive end-to-end gradient updates, limiting their scalability on large heterogeneous datasets.

\vspace{-3mm}
\subsection{Reinforcement Learning} 
\label{sec.rl}
\ac{RL} serves as a foundational paradigm for combinatorial optimization in AutoML~\cite{yangauto}. Policy Gradient methods such as PPO~\cite{schulman2017proximal} dominate domains characterized by low-latency simulators~\cite{phamefficient}. However, they are ill-suited for data preparation where the reward signal is both sparse and computationally expensive to obtain. Since these on-policy algorithms require fresh trajectories for every gradient update, their inherent low sample efficiency renders them computationally infeasible when each evaluation necessitates time-consuming downstream model training.

Consequently, value-based methods, particularly DQN~\cite{van2016deep} and its variants, have become the standard for modeling data preparation as a Markov Decision Process ~\cite{khurana2018feature}. These approaches are favored for their ability to leverage off-policy mechanisms, which enable the efficient reuse of historical experiences~\cite{chai}. Nevertheless, value maximization faces intrinsic limitations in this domain due to the delayed nature of supervision. As valid feedback is only received after executing the entire pipeline, \ac{TD} learning struggles to assign credit to early-stage decisions accurately~\cite{kumar2020conservative}. This ambiguity often drives agents to converge prematurely to suboptimal local maxima.

To address these issues, \ac{GFlowNets} offers a paradigm shift from reward maximization to reward sampling~\cite{bengio2021flow,bengio2023gflownet}. Unlike standard RL agents that seek a single optimal path, \ac{GFlowNets} treat the generation process as a flow-matching problem to sample trajectories proportional to their reward~\cite{silva2025gflownets}. 
This formulation theoretically ensures mode-covering exploration and effective credit assignment across long horizons without the prohibitive sampling costs of on-policy methods. While proven effective in scientific discovery tasks such as molecule generation~\cite{zhang2023let,jain2022biological,hu2023gflownet}, this flow-matching paradigm remains unexplored in automated data preparation. 
{\it Rather than directly applying it, FlowPipe tailors this architecture by instantiating a \ac{C-GFlowNets} conditioned on distilled semantic information. This structural adaptation injects a semantic prior into the flow logic, ensuring diverse exploration that strictly adheres to rigid logical dependencies.}
\vspace{-3mm}
\subsection{Foundation Models in Data Preparation}
Recent efforts to integrate Foundation Models into data preparation aim to bridge the semantic gap left by purely statistical approaches. 

\begin{table}[htbp]
\centering
\caption{Summary of Notations.}
\vspace{-2mm}
\label{tab:notations}
\small
\begin{tabular}{p{0.15\columnwidth} >{\centering\arraybackslash}p{0.8\columnwidth}}
\toprule
\textbf{Symbol} & \textbf{Description} \\ 
\midrule
$\mathcal{D}_{\text{train}}, \mathcal{D}_{\text{test}}$ & Training and test datasets \\
$\mathcal{T}$ & Set of component categories \\
$\mathcal{O}$ & Set of concrete operators \\
$\Omega$ & Hierarchical search space of pipelines \\
$P$ & Data preparation pipeline \\
$\pi$ & Policy \\
$\mathcal{M}$ & Downstream ML model \\
$\phi$ & Performance metric \\
$R$ & Reward derived from test performance \\
$\tau$ & Generation trajectory \\
\bottomrule
\end{tabular}
\end{table}

\textbf{Generative paradigms}~\cite{narayan2022can,zhang2023large} leverage \ac{LLM} to synthesize cleaning scripts directly via dialogue or few-shot prompting~\cite{li2024table}. For instance, ChatPipe~\cite{chen2024chatpipe} employs dialogue-driven interaction, while CatDB~\cite{fathollahzadeh2025catdb} utilizes refined metadata from data catalogs and prompt chaining to guide code generation. However, these models primarily function as probabilistic token generators guided by static rules rather than rigorous optimizers, resulting in operational hallucinations and execution failures. 

Alternatively, \textbf{representation-enhanced methods} such as CtxPipe~\cite{ctxpipe} utilize pre-trained embeddings (\ie, GTE-large~\cite{li2023towards}) to augment the state space of RL agents. Nevertheless, these systems typically rely on lossy data snapshots due to context limits~\cite{yin2020tabert} and employ shallow additive fusion that fails to capture the deep multiplicative dependencies between dataset semantics and operator logic. {\it Our work distinguishes itself by integrating LLM-enhanced priors via a rigorous \ac{GFlowNets} formulation with \ac{FiLM}-based modulation to achieve robust, semantic-aware pipeline synthesis.}

\section{Problem Definition}


In this section, we formalize the automated data preparation problem and introduce our formulation of pipeline synthesis as a generative flow process. Table~\ref{tab:notations} summarizes the frequently used notations.

To effectively navigate the combinatorial explosion of pipeline configurations, treating all available implementations w.r.t. all data preparation tasks as a flat list is inefficient. To address this, we leverage the inherent logical precedence in data preparation, where abstract functional primitives (\eg \textit{Imputation}) dictate the validity of specific algorithmic instantiations (\eg \texttt{Mean}). Accordingly, we structure the search space as a two-level hierarchy. This decouples functional intent from concrete implementation, analogous to the separation of a logical operator from its physical operators, enabling the model to prune vast suboptimal search regions by resolving high-level categories before committing to granular physical operator choices.

We formally define this structure as follows:
\begin{Def}[Hierarchical Operator Space]\label{def:space}
    Let $\mathcal{D}$ be the dataset. We define the search space $\Omega$ as a hierarchical collection of operations comprising two levels:
    \begin{itemize}[leftmargin=*]
        \item \textit{Component Categories} $\mathcal{T} = \{T_1, \dots, T_k\}$: These represent abstract functional primitives, such as \textit{Imputers} and \textit{Encoders}.
        \item \textit{Concrete Operators} $\mathcal{O}$: Each category $T_i$ maps to a specific subset of executable algorithms $\mathcal{O}_{T_i}$. For example, $\mathcal{O}_{\text{IM}} = \{\texttt{Mean}, \texttt{Median}, \allowbreak  \texttt{MostFreq}\}$.
    \end{itemize}
\end{Def}

Based on this hierarchy, a valid pipeline $P$ is formally defined as a linear sequence of executable operators instantiated from $\mathcal{O}$.
Concretely, $P$ manifests as a sequential transformation chain: the input data $\mathcal{D}$ is passed through a specific operator selected from $\mathcal{O}_{\text{IM}}$ (\eg \texttt{MeanImputer}) to handle missing values, and the transformed output is subsequently fed into an operator from $\mathcal{O}_{\text{EN}}$ (\eg \texttt{OneHotEncoder}) for feature encoding, strictly adhering to the sequential input-output dependencies defined by the operator logic.

To leverage the generative capabilities of \ac{GFlowNets}, we reformulate the static pipeline selection into a sequential decision-making process modeled as a flow network.

\begin{Def}[Generative Trajectory]
\label{def:trajectory}
We model the construction of a data preparation pipeline $P$ as a sequential trajectory $\tau = (s_0 \xrightarrow{a_0} s_1 \xrightarrow{a_1} \dots \xrightarrow{a_{f-1}} s_f)$.
Here, $s_0$ denotes the initial state, representing an empty pipeline. At each discrete time step $n$, the agent selects an action $a_n$, defined as appending an operator from the candidate set $\mathcal{O}$ that is valid under current constraints, to transition to a new state $s_{n+1}$ which encapsulates the updated pipeline topology. The process concludes at a terminal state $s_f$, corresponding to the fully constructed pipeline $P$. Each complete trajectory $\tau$ uniquely induces a pipeline $P(\tau)$.
\end{Def}

As depicted in the \textit{\ac{C-GFlowNets}} of Figure~\ref{fig:framework}, this trajectory generation is visualized as a directed path-finding process. The red dashed arrow (\eg $s_3 \to s_4$) highlights the active transition step where the agent samples the next operator based on the current policy.

\begin{Def}[Flow Consistency]
\label{def:flow_consistency}
    The core objective is to learn a policy that satisfies Flow Consistency: The \ac{GFlowNets} agent learns a stochastic policy $\pi$ to generate trajectories such that the trajectory likelihood $p(\tau)$ of sampling a complete trajectory is directly proportional to its performance reward $R(P)$:
    \begin{equation}
        p(\tau; \theta) \propto R(P; \mathcal{D}).
        \label{eq:flow_prop}
    \end{equation}
    This proportionality ensures that high-performing pipelines are sampled with higher probability. Unlike value-based RL formulations that primarily focus on optimizing a single trajectory, flow consistency distributes probability mass across multiple high-reward pipelines.
\end{Def}
Finally, based on the definitions above, we formally state the optimization problem.

Let $\mathcal{P}(\Omega)$ denote the set of valid pipeline sequences constructible from the hierarchical operator space $\Omega$.

\begin{figure*}[t]
    \begin{center}
    \includegraphics[width=0.97\textwidth]{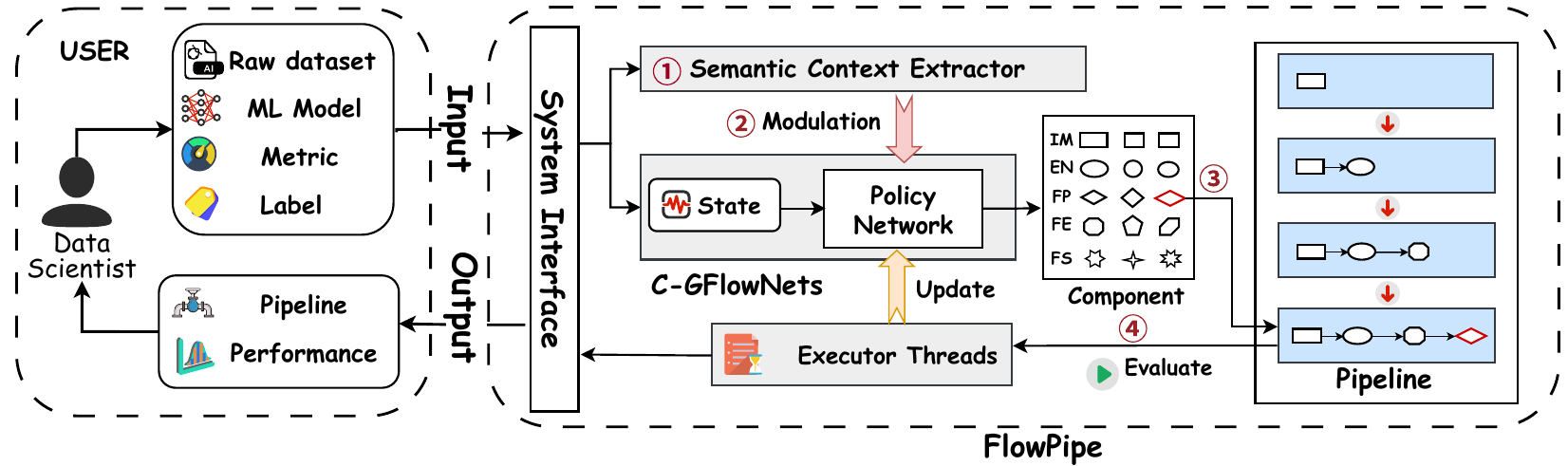}
    \vspace{-4mm}
    \caption{\textbf{The workflow of FlowPipe.} It illustrates the high-level interaction between the user and system, ranging from the initial request to the final pipeline output.}
    \label{fig:workflow} 
    \end{center}
    \vspace{-5mm}
\end{figure*}

\begin{Pro*}[Reward-Maximizing Pipeline Generation]
    Given a dataset $\mathcal{D}$ and a downstream model $\mathcal{M}$, we aim to learn a generative policy $\pi_\theta$ parameterized by $\theta$. 
    
    While the training phase aligns the generation probability with the reward distribution (Eq.~\eqref{eq:flow_prop}), the practical inference objective is to identify the optimal pipeline $P^*$. This entails finding the configuration that maximizes the validation reward $R$, approximated by sampling trajectories $\tau$ from the trained policy $\pi_\theta$:
    \begin{equation}
    P^* = \underset{P \in \mathcal{P}(\Omega)}{\text{argmax}}\ R(P; \mathcal{D})
    \approx \underset{\tau \sim \pi_\theta}{\text{argmax}}\ R(P(\tau); \mathcal{D}).
    \end{equation}
\end{Pro*}

\section{System Overview}
\label{sec:system_overview}

Figure~\ref{fig:workflow} depicts the architecture of \textbf{FlowPipe}. The system accepts a raw dataset $\mathcal{D}$ and a machine learning task $\mathcal{T}$ (comprising the target model, metric, and labels) as input, and synthesizes an executable data preparation pipeline $P$ to maximize downstream performance.

Unlike prior RL-based systems that rely on greedy action-value policies, FlowPipe synergizes semantic reasoning with generative flow modeling. By formulating pipeline construction as a probabilistic inference task via a \ac{C-GFlowNets}, and injecting global semantic priors through \ac{FiLM}-based modulation, FlowPipe steers the exploration toward semantically plausible and high-performing regions of the operator space. The detailed workflow proceeds as follows.

\noindent\textbf{Step 1 - Semantic Context Extraction.} 
The workflow commences at the \textit{System Interface}, which parses the input schema. To address the limitation where statistical profiles fail to capture operational logic, the \textit{Semantic Context Extractor} constructs a Structured Prompt and distills it into a dense modulation vector $\mathbf{V}_{sc}$ via an \ac{LLM}. 

\noindent\textit{Example:} Consider a used car dataset containing a ``Zip\_Code'' column. While standard statistics merely see a high-variance numerical distribution, our Extractor captures its semantic identity as a nominal geospatial identifier. This domain-aware insight allows the system to preemptively prune magnitude-based operations (e.g., Scaling), thereby significantly reducing the combinatorial search space by filtering out semantically incoherent trajectories. Details on the prompt construction and extraction strategy are in Section~\ref{sec:semantic}.

\noindent\textbf{Step 2 - Semantic-Modulated Policy Initialization.} 
Before the search phase, the Modulation vector is injected into the \textit{Policy Network} to guide the agent. We employ a \ac{FiLM} mechanism to recalibrate the policy's activations, effectively pruning operationally invalid actions based on the distilled semantics. 

\noindent \textit{Example:} For the ``Zip\_Code'' feature, the modulation signal suppresses the logits for Feature Preprocessing operators (\eg \texttt{Min Max Scaler}), as scaling a categorical code is semantically invalid. Conversely, it amplifies the probability of valid Encoder (EN) operators like \texttt{Label Encoder}. This narrows the search space and ensures logical correctness. The architectural implementation of this module is detailed in Section~\ref{sec:policy}.

\begin{figure*}
    \begin{center}
\includegraphics[width=0.97\textwidth]{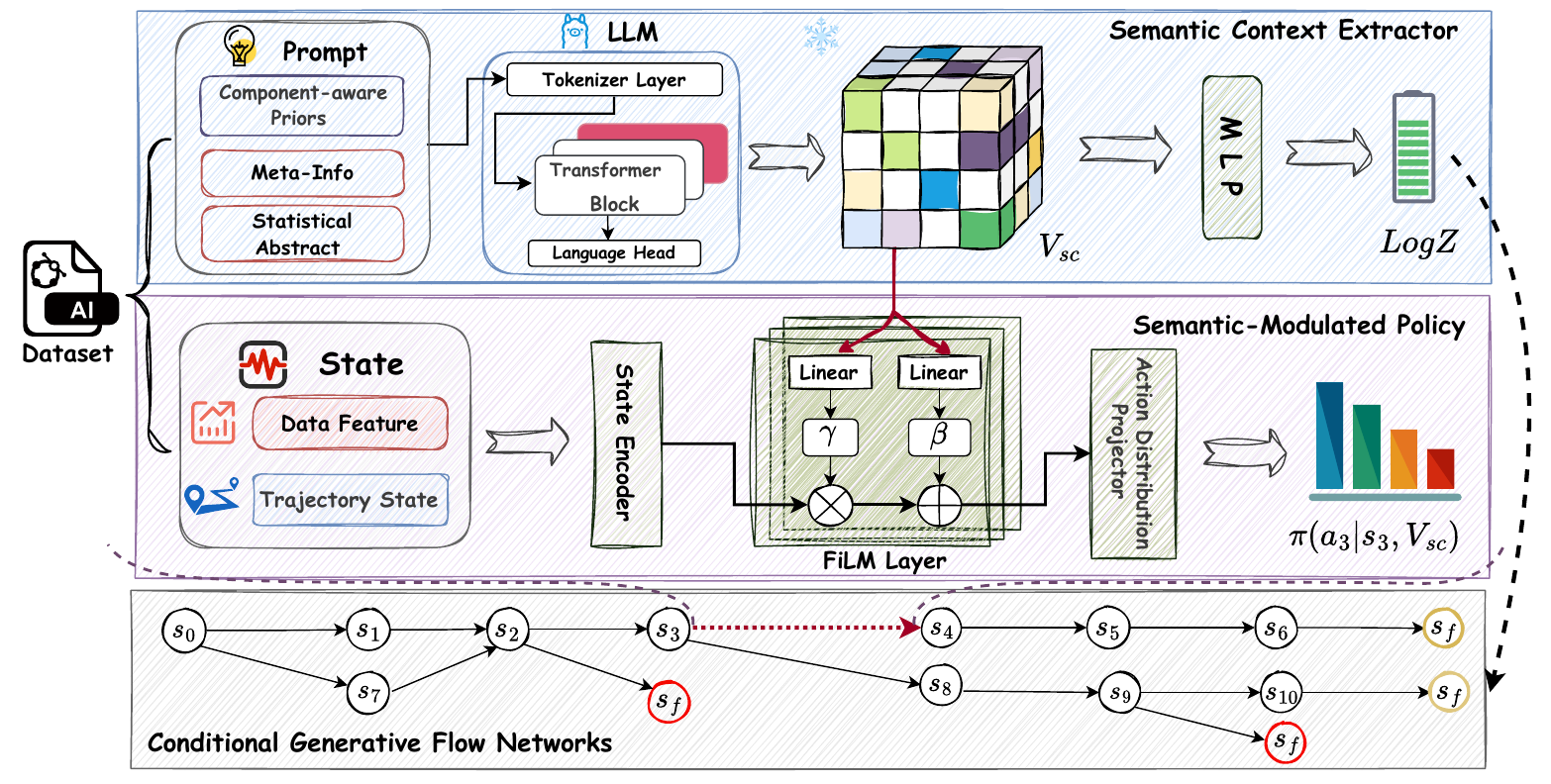}
\vspace{-3mm}
\caption{\textbf{Overview of the Decision-Making Process.} 
A Generative \ac{LLM} encodes the dataset into a semantic prior $\mathbf{V}_{sc}$ and normalization term $Log\mathbf{Z}$. 
This prior modulates the policy network via FiLM layers (middle) to condition the sequential pipeline synthesis (bottom), exemplified by the specific action sampling from $s_3$ to $s_4$ (red dashed lines).}
\label{fig:framework}
\end{center}
\vspace{-4 mm}
\end{figure*}

\noindent\textbf{Step 3 - Generative Pipeline Synthesis.} 
The \ac{C-GFlowNets} engine instantiates a sequential decision process, visualized as the state transition diagram at the bottom of Figure~\ref{fig:framework}. At each step, the \textit{Semantic-Modulated Policy} observes the current state and samples a specific component proportional to its learned flow probability. Specifically, the red dashed lines highlight an active decision instance where the policy evaluates state $s_3$ and samples an action to transition to $s_4$. We elaborate on the state representation and probabilistic modeling in Section~\ref{sec:gflownet}.

\noindent \textit{Example:} Guided by the policy, the agent samples a multi-hop transformation trajectory for ``Zip\_Code''. It first selects \texttt{Target Encoder}  to convert the high-cardinality categories into numerical risk scores, and subsequently chains \texttt{Robust Scaler} to normalize these scores. This sequential generation demonstrates FlowPipe's capability to build complex, topologically coherent workflows rather than isolated operator choices. We elaborate on the state representation and probabilistic modeling in Section~\ref{sec:gflownet}.

\noindent\textbf{Step 4 - Execution and Evaluation.}
Generated pipeline candidates are dispatched to \textit{Executor Threads}, which materialize transformations in an isolated runtime. The system trains the downstream machine learning model to compute a validation metric, serving as the terminal reward $R(\tau)$. This reward is used to update the policy via the \ac{TB}~\cite{niu2024gflownet} objective, which utilizes the normalization term $Log\mathbf{Z}$ passed from the Semantic Context Extractor (as indicated by the dashed arrow in Figure~\ref{fig:framework}). The detailed loss function is provided in Section~\ref{sec:training}.

In the subsequent sections, we detail the technical realization of these modules. Sec.~\ref{sec:semantic} elaborates on the \textit{Semantic Context Extractor} (Step 1), focusing on the LLM-driven derivation of semantic priors. Section~\ref{sec:gflownet} presents the \ac{C-GFlowNets} framework. Crucially, this framework serves as the unified engine underpinning Steps 2 -- 4: it integrates the semantic modulation mechanism (Step 2), drives the probabilistic synthesis process (Step 3), and governs the reward-based optimization via the Trajectory Balance objective (Step 4).

\vspace{-7mm}
\section{Semantic Context Extractor}
\label{sec:semantic}
As highlighted in Step 1 of the System Overview, standard statistical profiles often fail to capture the operational logic required for effective data preparation. Recent automated pipeline construction methods that incorporate embedding models, such as GTE~\cite{li2023towards}, typically generate local static representations derived from limited data snapshots including random row samples. Consequently, these approaches suffer from a significant lack of semantic information and fail to capture the global semantic context required to deduce operational logic.

To effectively distill this global semantic context, we construct the Generative Semantic Context Vector through a two-step process. We first transform the raw dataset into a coherent textual description via a Structured Semantic Prompt, encapsulating its intrinsic semantics and operational constraints. Subsequently, by processing this prompt through a Generative \ac{LLM} with a truncation-based extraction strategy, we distill the model's latent reasoning into a fixed dense embedding. This embedding bridges the semantic gap by aligning intrinsic data profiles with operational logical priors. Acting as a global conditional prior, it provides the necessary inductive bias to dynamically modulate the \ac{RL} state space and guide the \ac{GFlowNets} training process, ensuring the optimization prioritizes decisions that are both data-compatible and operationally valid.
\vspace{-3 mm}
\subsection{Structured Semantic Prompt}
\label{sec:prompt}
Constructing an effective semantic representation for large-scale tabular data presents a dual challenge. First, directly feeding raw datasets into \ac{LLM}s is computationally prohibitive due to finite context window constraints. Second, heuristic approximations adopted by baselines like CtxPipe, which rely on randomly subsampling small fragments, suffer from severe \textit{local sampling bias}. Consequently, these methods fail to capture critical global properties including class imbalance and long-tail distributions. Furthermore, raw data ingestion lacks the necessary task-specific inductive bias. Without explicit instructions, the \ac{LLM} remains agnostic to the underlying optimization objective. To overcome these limitations, we explicitly construct a structured prompt $\mathcal{P}$ that functions as a mechanism for \textit{lossless semantic compression}. 
\reva{To ensure scalability, this architecture explicitly excludes raw data rows, completely decoupling token complexity from dataset volume. Specifically, by relying on schema-level metadata and dataset statistics summaries, scaling to millions of instances introduces zero additional tokens, while the footprint of increasing feature dimensionality remains strictly marginal.}
Specifically, this prompt is composed of three functional blocks designed to concretize intrinsic semantics and operational constraints.

First, the \textbf{Meta-Info} block functions as the semantic anchor of the prompt. It aggregates explicit metadata, including the dataset name, attribute headers, and basic task properties, to construct a high-level identity profile of the data. These textual identifiers, such as column names like ``credit\_score'', encapsulate rich semantic cues that enable the \ac{LLM} to infer domain-specific strategies that escape shallow statistical analysis. For instance, distinguishing ``systolic\_bp'' from ``credit\_score'' helps the model prioritize medical interpretability over financial noise tolerance.
\reva{Critically, since this block targets only schema-level headers, its token count remains invariant to row volume. Increasing dimensionality ($D$) introduces only marginal overhead, as each additional feature merely appends a single textual identifier (\eg the column name) to the sequence.}

Complementing this anchor, the \textbf{Statistical Abstract} synthesizes a holistic distributional profile. It translates discrete numerical properties into a serialized schema narrative. Designed as a compact set of critical descriptors, this block captures the topological features defining dataset quality. Specifically, it enumerates per-column statistics including central tendency metrics like mean and mode, dispersion indicators such as range, and missingness ratios to expose structural constraints. These signals allow the model to assess dataset density for imputation and arbitrate between encoding strategies based on cardinality. Moreover, it explicitly maps strong pairwise correlations where the absolute Pearson coefficient $|\rho|$ exceeds 0.5, revealing intrinsic feature space redundancy to guide feature selection. 
\reva{Likewise, our design only uses the statistics collected from the dataset rather than involving raw data instances. Therefore, this block's length is completely unaffected by row volume, requiring only a concise statistical profile for each added feature.}

Crucially, we inject \textbf{Component-aware Priors} into the prompt to bridge the semantic gap between static statistical profiles and dynamic operator selection. This block functions as a logical instruction set that explicitly maps serialized data characteristics to operator preconditions. By linguistically articulating applicability heuristics, such as specifying that significant feature redundancy warrants PCA and high-cardinality categories necessitate Target Encoding, we establish a basis for the subsequent \ac{LLM} encoding phase. This formulation strictly constructs the logical context to guide the downstream reasoning, ensuring that the latent representation generated later is constrained by operational validity. 
\reva{Importantly, the information injected into this block is entirely independent of the dataset's volume and dimensionality. Therefore, its token footprint remains constant, ensuring scalability with zero additional overhead as the dataset grows.}



\vspace{-4mm}
\subsection{LLM-Enhanced Latent Reasoning}
\label{sec:llm_reasoing}
Given the structured prompt $\mathcal{P}$, our objective is to distill the \ac{LLM}'s latent reasoning regarding the alignment of intrinsic data properties with operational prerequisites into a dense, operable vector.

Standard discriminative embedding models like BERT~\cite{bert} or GTE~\cite{li2023towards} are constrained by optimization objectives that prioritize static lexical similarity. While effective for matching semantically related terms, they lack the causal reasoning mechanism required to process conditional logic. For instance, GTE may relate ``skewness'' to ``distribution'' but fails to infer that high skewness necessitates Log-Transformation. To address this limitation, we leverage the context-adaptive reasoning capacity of the Generative \ac{LLM} to preserve the continuous reasoning state implicit in the model activations rather than collapsing it into discrete tokens.

Structurally, the Generative LLM follows a standard decoder-only transformer architecture comprising a stack of $K$ blocks followed by a Language Model head. We forward-propagate the prompt $\mathcal{P}$ through the entire transformer stack but truncate the execution explicitly at the output of the $K$-th block to bypass the LM head. Instead of projecting the hidden states into the high-dimensional vocabulary space for token generation, we extract the raw output features from the final transformer layer, denoted as $\mathbf{H} \in \mathbb{R}^{L \times d}$ for sequence length $L$ and hidden dimension $d$. 
We deliberately employ open-weights models to ensure white-box access to these internal representations. This design choice is critical for system deployment: it allows FlowPipe to bypass the high-latency iterative decoding process typical of black-box APIs, while ensuring data privacy by performing all semantic reasoning within the local runtime environment.

\reva{To resolve the variable sequence length driven by the feature dimensionality $D$, we apply mean pooling over the token dimension of $\mathbf{H}$. This operation compresses the high-dimensional semantic interactions into a compact, fixed-size manifold $\mathbf{V}_{sc}$. Crucially, producing a constant-sized representation decouples the LLM's verbosity from the downstream policy, natively preventing the curse of dimensionality on wide tabular datasets.} Computationally, this design strictly separates heavy semantic reasoning from efficient structure search. Furthermore, $\mathbf{V}_{sc}$ is generated and cached offline as a static global prior, amortizing the LLM's inference cost and ensuring the iterative online search remains lightweight without recurrent overhead.
\vspace{-1mm}
\section{Conditional Generative Flow Networks}
\label{sec:gflownet}

Having distilled the intrinsic data semantics into the modulation vector $\mathbf{V}_{sc}$, we now introduce the generative engine that leverages this prior. The core generative engine of FlowPipe is instantiated as a \ac{C-GFlowNets}. Unlike standard RL agents that focus on optimizing a single high-reward trajectory,  \ac{C-GFlowNets} samples pipelines proportional to their reward to ensure diverse exploration of high-performance configurations.
\vspace{-1mm}
\subsection{Probabilistic Modeling Instantiation}
Building upon the generative trajectory formulation in Definition~\ref{def:trajectory}, we instantiate the pipeline construction process as a linear sequential decision-making task, explicitly conditioned on the global semantic context $\mathbf{V}_{sc}$. In FlowPipe, the \ac{C-GFlowNets} is conditioned on a global semantic vector $\mathbf{V}_{sc}$, which captures dataset-specific constraints and guides the generation of valid pipeline trajectories.

Formally, we parametrize a policy $\pi(s_{n+1}|s_n, \mathbf{V}_{sc}; \theta)$ with learnable parameters $\theta$. The core objective is to learn a policy that satisfies the Flow Consistency condition, where the marginal likelihood $p(\tau | \mathbf{V}_{sc})$ of generating a complete trajectory $\tau$ is directly proportional to its terminal reward $R(\tau)$:
\begin{equation}
    p(\tau \mid \mathbf{V}_{sc}) \propto R(\tau).
\end{equation}

This unified probabilistic formulation mitigates the structural dissonance inherent in fragmented Multi-DQN architectures, thereby improving the credit assignment problem. By treating the generation process as a coherent flow, it compels the agent to internalize the global dependency between early-stage decisions (e.g., imputation) and downstream outcomes (e.g., classification accuracy) within a single policy $\pi$. Furthermore, conditioning the entire flow on $\mathbf{V}_{sc}$ transforms the optimization from a blind combinatorial search into a semantic-aware probabilistic sampling process, effectively constraining exploration to the high-probability manifold defined by the data's intrinsic properties. 

\vspace{-2 mm}
\subsection{State Representation}
\label{sec.state}
A robust state representation is critical for accurate environmental perception. At each decision step $n$, the agent observes a composite state $s_{n}$ that encapsulates both the sequential history of the pipeline and the evolving statistical properties of the data. As shown in the State Block of Figure~\ref{fig:framework}, we construct $s_{n}$ via the concatenation of two distinct feature vectors $s_{n} = [\mathbf{s}^{tr}_{n} \oplus \mathbf{s}^{d}_{n}]$. 
\revc{Crucially, because each applied operator immediately mutates the underlying dataset and extends the pipeline history, both $\mathbf{s}^{tr}_{n}$ and $\mathbf{s}^{d}_{n}$ are dynamically recomputed at every transition step to ensure the policy evaluates strictly up-to-date environmental feedback.}

\noindent\textbf{Trajectory State} ($\mathbf{s}^{tr}_{n}$).
This vector encodes the \textbf{sequential context} of the partial pipeline.
To avoid the computational latency associated with recurrent sequence modelers, such as LSTMs, we implement an efficient \textit{Explicit Sequential Encoding}.
We map the current pipeline configuration to a fixed-length binary vector $\mathbf{s}^{tr}_{n} \in \{0, 1\}^{|\mathcal{T}|}$, where each dimension functions as an indicator for a specific operator category.
For instance, given a vocabulary ordered as $\{\texttt{Imputer}, \texttt{Encoder}, \dots\}$, a pipeline that has already applied an imputer is represented as $[1, 0, \dots]$.
This lightweight representation provides an instantaneous structural snapshot, enabling the policy to strictly enforce sequential constraints, such as prohibiting redundant imputation, via low-cost logical masking.

\textbf{Data Feature State} ($\mathbf{s}^{d}_{n}$).
This vector captures the granular statistical profile of the intermediate dataset transformed by the previous $n$ operators. It includes meta-features such as skewness, kurtosis, sparsity, and correlation coefficients. A key challenge is the vast numerical range of these statistics where cardinality can range from 2 to $10^6$ while variance remains infinitesimal. Feeding raw values directly into neural networks often leads to gradient instability and poor convergence. To mitigate this, we utilize a {\it Dual-Channel Numerical Embedding}. We decompose each continuous meta-feature $\psi$ into a tuple representing its magnitude and precision via the mapping $\psi \rightarrow (\text{frac}(\psi), \exp(\psi))$. This representation effectively disentangles numerical scale from value distribution to prevent information collapse for outlier values and ensure stable training across heterogeneous datasets.
\vspace{-1mm}
\subsection{Semantic-Modulated Policy Network}
\label{sec:policy}
The Semantic-Modulated Policy serves as the decision-making kernel. It maps the composite state $s_n$ to a probability distribution over the operator space $\mathcal{O}$.

Context-aware baselines such as CtxPipe rely on \textit{shallow additive fusion}. As highlighted by the problem of semantic detachment, this formulation functions merely as an \textit{extrinsic bias term}, which is structurally insufficient to capture the deep, multiplicative dependencies between data profiles and operator logic. To resolve this, we employ \ac{FiLM} for deep semantic modulation.
\revb{To resolve this, we adopt Feature-wise Linear Modulation (\ac{FiLM})~\cite{perez2018film}. While recognized as an established neural conditioning technique, we exploit its structural properties specifically for deep semantic modulation in data preparation.} 
By utilizing $\mathbf{V}_{sc}$ to multiplicatively recalibrate feature activations, this mechanism functions as a differentiable logic gate. It enables dynamic logic adaptation and performs the soft action suppression essential for mitigating \textit{exploration inefficiency}.

\textbf{FiLM-Based Semantic Modulation.}
We implement semantic injection via a \ac{FiLM} mechanism. We first project the raw state $s_t$ into a latent feature sequence $h_{t}$. Simultaneously, the invariant context $\mathbf{V}_{sc}$ is mapped to layer-specific affine parameters via a projection head $g^{(l)}$. For the $l$-th layer of the policy network, the feature modulation is defined as:
\begin{equation}
    h_{n}^{l} = \text{FiLM}(h_{n}^{l-1} \mid \mathbf{V}_{sc}) = \gamma^{l} \odot h_{n}^{l-1} + \beta^{l},
\end{equation}
here $\odot$ denotes the element-wise Hadamard product. The projection head $g^{(l)}$ consists of two separate multilayer perceptrons that predict the scaling vector $\gamma^l$ and shifting vector $\beta^l$ respectively.

This design enables a dual-mechanism control over the decision logic. The multiplicative coefficient $\gamma^{l}(\mathbf{V}_{sc})$ functions as a dynamic feature re-calibration mechanism. It selectively amplifies or suppresses specific feature channels based on the logical priors encoded in $\mathbf{V}_{sc}$. Conversely, the additive coefficient $\beta^{l}$ introduces a context-dependent bias shift, adjusting the activation thresholds of operators to align with data properties. By applying this modulation recursively across all $L$ layers of the policy network, FlowPipe achieves hierarchical semantic alignment. Early layers effectively prune low-level incompatible signals while deeper layers refine abstract operator sequencing strategies.

Crucially, this mechanism implements the \textit{Soft Action Suppression} required to mitigate exploration inefficiency.
Revisiting the ``Zip\_Code'' case where $\mathbf{V}_{sc}$ encodes a nominal constraint, the projection head learns to predict near-zero $\gamma$ coefficients for magnitude-dependent operators, such as \texttt{MinMaxScaler}.
This operation effectively suppresses forward signal propagation for these semantically incompatible actions and significantly reduces their gradient contributions.
Consequently, the agent is guided to explore valid regions of the operator space, which reduces exploration variance and accelerates convergence compared to unconstrained exploration in standard RL.

\textbf{Action Distribution Modeling.}
The semantically modulated feature vector is passed to an {\it Action Distribution Projector} which maps the latent representation to logits $\tilde{L}_n$ over the operator space $\mathcal{O}$. 
In contrast to value-based baselines (e.g., DQN) that often induce near-deterministic behavior via greedy action selection and to support robust pipeline discovery under a large combinatorial search space, we model the action choice as a Categorical Distribution:
\begin{equation}
    a_{n} \sim \pi(a_{n} = k | s_n, V_{sc}) = \text{Softmax}(\tilde{L}_n) = \frac{\exp(\tilde{L}_{n}^{k})}{\sum_{j \in \mathcal{O}} \exp(\tilde{L}_{n}^{j})},
\end{equation}
where $k$ represents the operator chosen by the policy at step $n$, while $j$ enumerates all candidate operators in $\mathcal{O}$.  This stochastic sampling mechanism prevents premature mode collapse and supports the discovery of multiple high-quality pipeline configurations.

\vspace{-1mm}
\subsection{Training Objective}
\label{sec:training}

Training faces delayed, expensive end-to-end evaluation feedback. We thus adopt the \ac{TB} objective for Conditional GFlowNets, where the semantic context $\mathbf{V}_{sc}$ conditions the trajectory distribution via a partition function.

\noindent\textbf{Trajectory-Level Supervision for Credit Attribution.}
Value-based RL methods typically rely on step-wise bootstrapping, which often propagates delayed validation signals inefficiently over long pipeline horizons, making it difficult to attribute final performance to early-stage decisions.
In contrast, TB-style flow matching enforces trajectory-level consistency and provides holistic supervision that relates the terminal reward $R(\tau)$ to all decisions along the trajectory.
As a result, the objective induces an implicit gradient path from the terminal validation reward back to early pipeline decisions, without relying on step-wise bootstrapping.

Formally, following the Conditional GFlowNet formulation, we obtain the context-dependent partition function $\log\mathbf{Z}$ by mapping the semantic condition vector $\mathbf{V}_{sc}$ through a lightweight MLP. 
In addition, because the combinatorial search space contains many invalid or incomplete pipelines that yield no reward, we introduce a \textit{failure-aware weighting} scheme to limit their influence during optimization. 
Specifically, we minimize the following TB-style loss:
\begin{equation}
    \mathcal{L}(\tau; \theta) = w(\tau) \cdot \left(
    \log\mathbf{Z}
    + \sum_{n=0}^{|\tau|-1} \log \pi(a_{n} \mid s_n; \mathbf{V}_{sc})
    - \log R(\tau)
    \right)^2 .
\end{equation}

This objective combines three complementary components.
First, $\log \mathbf{Z}$ provides context-conditioned normalization, which stabilizes learning across datasets with heterogeneous reward scales.
Second, the trajectory-level log-probability term distributes supervision across all pipeline decisions, improving credit attribution with respect to the final validation reward.
Finally, $w(\tau)$ implements failure-aware weighting: it assigns unit weight to valid pipelines and a reduced penalty factor $\lambda \in [0,1)$ to invalid or incomplete executions, thereby focusing learning on valid, high-quality pipeline configurations.
All parameters $\theta$ are jointly optimized by minimizing this flow-matching loss.

\vspace{-1mm}
\begin{table}[ht]
\centering
\small 
\caption{Selected components and their types.}
\vspace{-3mm}
\begin{tabular}{@{}p{0.32\linewidth} p{0.63\linewidth}@{}} 
\toprule
\textbf{Component \newline categories} & \textbf{Operators} \\
\midrule
Imputer \textbf{(IM)} & Mean, Median, Most Frequent \\ \addlinespace
Encoder \textbf{(EN)} & Numeric Data, Label Encoder, One-hot Encoder \\ \addlinespace
Feature Preprocessing \textbf{(FP)} & Min Max Scaler, Max Absolute Scaler, Robust Scaler, Standard Scaler, Quantile Transformer, Normalizer, Power Transformer, K-bins Discretizer \\ \addlinespace
Feature Engineering \textbf{(FE)} & Polynomial Features, Interaction Features, PCA, Kernel PCA, Incremental PCA, Truncated SVD, Random Trees Embedding \\ \addlinespace
Feature Selection \textbf{(FS)} & Variance Threshold \\
\bottomrule
\end{tabular}
\label{tab:components}
\end{table}
\vspace{-2mm}
\section{Experiment}
\label{sec.exp}
Our experiments target answering the following four questions:

\begin{itemize}[leftmargin=*]
    \item \textbf{RQ1 (Effectiveness):} How does the predictive quality of pipelines discovered by FlowPipe compare to \ac{SOTA} baselines across diverse real-world datasets?
    
    \item \textbf{RQ2 (Efficiency):} How do FlowPipe's offline training costs and online inference latency compare to traditional search-based and meta-learning approaches?
    
    \item \textbf{RQ3 (Component Analysis):} What is the contribution of each core component—specifically the LLM-driven semantic reasoning and FiLM modulation—to the overall performance?
    
    \item \textbf{RQ4 (Scalability \& Robustness):} How does the framework scale with increasing dataset sizes, and is the performance robust to hyperparameter variations?
\end{itemize}
\vspace{-2mm}
\subsection{Experimental Setup}
\label{sec.setup}
\textbf{Hardware and OS.} Experiments are conducted on an Ubuntu 22.04 server with dual AMD EPYC 7402 CPUs (48 cores), 256GB RAM, and NVIDIA RTX 4090 GPUs (CUDA 12.2). For SAGA, the Java heap size is set to 80GB.

\textbf{Implementation Details.}
For the semantic context extraction module, we employ \textbf{Llama-3.1-8B-Instruct}~\cite{grattafiori2024llama} as the backbone Large Language Model. The model is frozen during the GFlowNet training phase, serving solely as a static feature extractor to distill reasoning-rich semantic vectors. All neural network components within the GFlowNet agent are implemented using PyTorch.


\textbf{Datasets.} We use a comprehensive suite of real-world datasets sourced from OpenML~\cite{vanschoren2014openml}, UCI~\cite{asuncion2007uci}, and Kaggle. To comprehensively evaluate scalability and out-of-distribution generalization, 
\reva{these datasets encompass a diverse range of feature dimensionalities ($D \in [4, 343]$) and are strictly separated into disjoint sets.} We train our agent on the HAIPipe dataset collection~\cite{haipipe} (353 datasets) and perform zero-shot evaluation on the DiffPrep~\cite{diffprep} (18 datasets) and DeepLine~\cite{deepline} (56 datasets) collections. We ensure zero overlap between the training and test collections to verify the model's capability to synthesize pipelines for entirely unseen tasks.

\textbf{Components.}
To ensure broad compatibility, all pipeline components adopt the scikit-learn API, encapsulating each operator as a class with \texttt{fit()} and \texttt{transform()} methods. This inherently supports User-Defined Functions (UDFs), allowing users to seamlessly inject custom logic. As detailed in Table~\ref{tab:components}, our default search space provides a comprehensive taxonomy consistent with \ac{SOTA} benchmarks~\cite{haipipe,ctxpipe}. 
\revb{Beyond this default formulation, our architecture natively supports the integration of broader data preparation operations detailed in extensive categorizations~\cite{hameed2020data, mozzillo2023evaluation}. This design facilitates the continuous expansion of both \textit{macro-categories} and \textit{micro-operators} without triggering a combinatorial explosion.} Topologically, we employ a flexible linear architecture where the agent dynamically selects, reorders, or bypasses components entirely, enabling the discovery of diverse, non-trivial operator sequences.

\textbf{Downstream Model.}
Adhering to standard evaluation protocols in automated data preparation~\cite{ctxpipe}, we utilize Logistic Regression as the downstream evaluator.
This choice ensures rigorous baseline comparison and guarantees broad reproducibility, as the model is natively supported by major frameworks including PyTorch, TensorFlow, scikit-learn, and Apache SystemDS~\cite{boehmsystemds}.

\begin{table*}[htbp]
  \centering
\caption{Performance on DiffPrep dataset collection. The \textbf{Improvement} row denotes FlowPipe's gain over \ac{SOTA}.}
\vspace{-2mm}
  \begingroup
  \small
  \renewcommand{\arraystretch}{1}
  \setlength{\aboverulesep}{0.3ex}
  \setlength{\belowrulesep}{0.3ex}
  \makebox[\textwidth][c]{
    \scalebox{1.02}{
      \begin{tabular}{lcccccc}
        \toprule
        & \multicolumn{6}{c}{\textbf{DiffPrep Datasets Collections}} \\
        \cmidrule(lr){2-7}
        \textbf{Setup} & \textbf{Test Accuracy} $\uparrow$ & \textbf{Ranking} $\downarrow$ & \textbf{Precision} $\uparrow$ & \textbf{Recall} $\uparrow$ & \textbf{F1-score} $\uparrow$ & \textbf{Inference Time (s)} $\downarrow$ \\
        \midrule
        DEF    (\textit{SIGMOD'23})           & 0.724 & 9.512 & 0.605 & 0.629 & 0.604 & 36.237 \\
        RS     (\textit{SIGMOD'23})            & 0.761 & 8.125 & 0.647 & 0.673 & 0.650 & 803.888 \\
        DP-Fix   (\textit{SIGMOD'23})      & 0.780 & 6.422 & 0.681 & 0.727 & 0.687 & 1024.845 \\
        DP-Flex  (\textit{SIGMOD'23})       & 0.784 & 4.188 & 0.697 & 0.722 & 0.695 & 11148.911 \\
        DL  (\textit{SIGKDD'20})             & 0.704 & 11.115 & 0.616 & 0.584 & 0.570 & \underline{18.923} \\ 
        HAI-AI  (\textit{SIGMOD'23})        & 0.760 & 8.194 & 0.705 & 0.619 & 0.613 & 22.195 \\
        SAGA (\textit{SIGMOD'23})          & 0.731 & 8.286 & 0.647 & 0.640 & 0.618 & 1384.824 \\
        \revc{TPE (\textit{arXiv'23})} & \revc{0.781} & \revc{4.580} & \revc{0.633} & \revc{0.644} & \revc{0.637} & \revc{1299.110} \\ 
        \reva{Direct Prompting}        & \reva{0.712} & \reva{10.420} & \reva{0.623} & \reva{0.631} & \reva{0.615} & \reva{142.508} \\ 
        \reva{ChatPipe (\textit{SIGMOD'24})} & \reva{0.783} & \reva{4.310} & \reva{0.751} & \reva{0.718} & \reva{0.732} & \reva{278.441} \\  
        CtxPipe (\textit{SIGMOD'24})       & \underline{0.806} & \underline{3.655} & \underline{0.774} & \underline{0.736} & \underline{0.745} & 65.203 \\
        SwiftDP (\textit{ICDE'25})       & 0.728 & 9.182 & 0.700 & 0.702 & 0.683 & \textbf{10.203} \\ 
        \textbf{FlowPipe} \textit{(Ours)} & \textbf{0.896} & \textbf{1.139} & \textbf{0.788} & \textbf{0.794} & \textbf{0.780} & 51.516 \\
        \midrule
        \textbf{Improvement} & \textbf{+11.17\%} & \textbf{+2.516} & \textbf{+1.81\%} & \textbf{+7.88\%} & \textbf{+4.70\%} & -- \\
        \bottomrule
      \end{tabular}
    }
  } 
    \endgroup
\label{tab5.1:diff_avg}
\vspace{-2mm}
\end{table*}

\begin{table*}[htbp]
  \centering
\caption{Performance on DeepLine dataset collection. The \textbf{Improvement} row denotes FlowPipe's gain over \ac{SOTA}.}
\vspace{-2mm}
  \begingroup
  \small
  \renewcommand{\arraystretch}{1}
  \setlength{\aboverulesep}{0.3ex}
  \setlength{\belowrulesep}{0.3ex}
  \makebox[\textwidth][c]{
    \scalebox{1}{
      \begin{tabular}{lcccccc}
        \toprule
        & \multicolumn{6}{c}{\textbf{DeepLine Datasets Collections}} \\
        \cmidrule(lr){2-7}
        \textbf{Setup} & \textbf{Test Accuracy} $\uparrow$ & \textbf{Ranking} $\downarrow$ & \textbf{Precision} $\uparrow$ & \textbf{Recall} $\uparrow$ & \textbf{F1-score} $\uparrow$ & \textbf{Inference Time (s)} $\downarrow$ \\
        \midrule
        DEF (\textit{SIGMOD'23})            & 0.760 & 8.722 & 0.754 & 0.769 & 0.751 & 268.899 \\
        RS    (\textit{SIGMOD'23})            & 0.776 & 8.328 & 0.755 & 0.776 & 0.751 & 2069.028 \\
        DP-Fix   (\textit{SIGMOD'23})        & 0.780 & 7.356 & 0.767 & 0.780 & 0.764 & 325.648 \\
        DP-Flex  (\textit{SIGMOD'23})       & 0.789 & 6.122 & 0.772 & 0.789 & 0.769 & 2183.917 \\
        DL         (\textit{SIGKDD'20})     & 0.740 & 9.921 & 0.738 & 0.740 & 0.727 & 48.439 \\
        HAI-AI    (\textit{SIGMOD'23})     & 0.801 & 5.189 & 0.781 & 0.753 & 0.767 & 16.780 \\
        SAGA      (\textit{SIGMOD'23})     & 0.809 & \underline{4.132} & 0.714 & 0.731 & 0.723 & 784.132 \\
        \revc{TPE (\textit{arXiv'23})} & \revc{0.792} & \revc{5.850} & \revc{0.704} & \revc{0.711} & \revc{0.708} & \revc{382.341} \\ 
        \reva{Direct Prompting}        & \reva{0.723} & \reva{11.450} & \reva{0.619} & \reva{0.611} & \reva{0.615} & \reva{112.107} \\ 
        \reva{ChatPipe (\textit{SIGMOD'24})} & \reva{0.777} & \reva{7.620} & \reva{0.741} & \reva{0.718} & \reva{0.730} & \reva{284.411} \\  
        CtxPipe    (\textit{SIGMOD'24})    & \underline{0.813} & 5.611 & \underline{0.784} & \underline{0.787} & \underline{0.775} & 14.310 \\
        SwiftDP      (\textit{ICDE'25})    & 0.780 & 6.845 & 0.692 & 0.636 & 0.643 & \textbf{5.760} \\
        \textbf{FlowPipe} \textit{(Ours)} & \textbf{0.912} & \textbf{1.089} & \textbf{0.813} & \textbf{0.823} & \textbf{0.805} & \underline{10.510} \\
        \midrule
        \textbf{Improvement} & \textbf{+12.17\%} & \textbf{+3.043} & \textbf{+3.70\%} & \textbf{+4.57\%} & \textbf{+3.87\%} & \textbf{--} \\ 
        \bottomrule
      \end{tabular}
    }
  } 
    \endgroup

  \label{tab5.1:deepline_avg}
  \vspace{-2mm}
\end{table*}

\textbf{Evaluation Metrics.}
We adopt Test Accuracy as the primary metric $\phi$ to align with baseline protocols, supplemented by Macro-Precision, Recall, and F1-score. We select macro-averaging because micro-averaged metrics are mathematically equivalent to accuracy in our setup. To assess efficiency, we report Pipeline Runtime, defined as the end-to-end wall-clock time comprising pipeline construction, execution, and downstream model training. This metric is critical for the evolving data analytics lifecycle, where distribution shifts~\cite{lee2017big,luo2021mlcask} necessitate frequent pipeline reconstruction.

\vspace{-2mm}
\subsection{Performance Comparison}
\label{sec:baseline}
Tables~\ref{tab5.1:diff_avg} and \ref{tab5.1:deepline_avg} present the \textbf{average performance} comparison results against baselines on the DiffPrep and DeepLine dataset collection, respectively. Detailed performance breakdowns for individual datasets are provided in the Appendix.
\vspace{-2mm}
\subsubsection{Baselines} We compare FlowPipe against the following \ac{SOTA} methods. To ensure a fair comparison, we adopt their open-source implementations and strictly adhere to the hyperparameter configurations recommended in their original papers or documentation:
\begin{itemize}[leftmargin=*]
\vspace{-8mm}
\item \textbf{Default (DEF)~\cite{diffprep}:} A standard rule-based pipeline aligned with commercial AutoML frameworks like Azure~\cite{kang2024azure} and H2O~\cite{ledell2020h2o}. It performs mean and mode imputation for numerical and categorical features, respectively, followed by standardization.

\item \textbf{Random Search (RS)~\cite{diffprep}:} Randomly instantiates 20 pipelines under the DP-Fix template, reporting the best-validated one.

\item \textbf{DiffPrep-Fix (DP-Fix) and DiffPrep-Flex (DP-Flex)~\cite{diffprep}:} We use the configurations reported in~\cite{diffprep}. DP-Fix searches under a fixed logical pipeline structure, while DP-Flex performs gradient-based optimization over a continuous relaxation to enable flexible component selection.
\item \textbf{DeepLine (DL)~\cite{deepline}:} A monolithic DQN baseline that uses a single Q-network for end-to-end decision making~\cite{deepline}. We use the authors' implementation with the configuration in~\cite{deepline} and train for 150{,}000 steps, following authors' recommendation.
\item \textbf{HAIPipe (HAI-AI)~\cite{haipipe}:} The AI-pipe agent from HAIPipe~\cite{haipipe}, built on a Multi-DQN architecture with stage-wise Q-networks for sequential pipeline generation. We use the authors' released model trained for 56{,}000 steps.
\item \textbf{SAGA~\cite{saga}:} An evolutionary baseline using genetic programming for data cleaning pipeline optimization, configured as in~\cite{saga}.
\item \revc{\textbf{TPE}~\cite{tpe}: A \ac{SOTA} Bayesian optimization algorithm using density estimation and likelihood ratio for efficient black-box search.}
\item \reva{\textbf{Direct Prompting}: A zero-shot baseline  where Llama-3.1-8B generates pipelines  directly from FlowPipe's semantic context.} 
\item \reva{\textbf{ChatPipe~\cite{chen2024chatpipe}:} A conversational LLM baseline that generates pipelines for automated data preparation.}
\item \textbf{CtxPipe~\cite{ctxpipe}:} A context-aware RL baseline that leverages pretrained language models to encode dataset semantics, and adopts a \emph{Multi-DQN} architecture~\cite{ctxpipe}.
\item \textbf{SwiftDP~\cite{swiftdp}:} An MCTS-based framework that combines meta-learning and attention to accelerate pipeline generation~\cite{swiftdp}.
\end{itemize}

\vspace{-3mm}
\subsubsection{Analysis of Effectiveness (RQ1)}
\label{sec:exp_performance}
We first evaluate the performance of constructed pipelines on test datasets for all setups. This is measured by the evaluation metrics $\phi$ and its ranking across setups per dataset.

\textbf{Test Accuracy.}
Tables~\ref{tab5.1:diff_avg} and \ref{tab5.1:deepline_avg} summarize aggregated performance, with dataset-level breakdowns deferred to the \textbf{Appendix}.
\revc{Evaluated rigorously across 10 random seeds, FlowPipe exhibits exceptional systemic stability with a negligible accuracy variance of strictly $\pm$0.003, robustly outperforming deterministic baselines.}

In aggregate, FlowPipe leads significantly with average accuracies of \textbf{0.896} on the DiffPrep collection and \textbf{0.912} on the DeepLine collection, surpassing the previous SOTA results of 0.806 and 0.813, respectively. These improvements are substantial in the context of data preparation~\cite{deepline,diffprep}. FlowPipe’s integration of contextual information results in significant performance gains, particularly on complex tasks. For instance, the \textit{Accident} dataset shows a 41.3\% relative improvement in accuracy compared to the best baseline ($0.978$ vs $0.692$). Datasets such as \textit{abalone}, \textit{connect-4}, and \textit{micro} also see notable improvements, benefiting from high-dimensional feature interactions or semantic dependencies, which FlowPipe’s LLM-driven context plug-in effectively captures to optimize performance.

\textbf{Optimality Gap Analysis.}
We compare FlowPipe with an approximated exhaustive search ($ES^*$) utilizing 10,000 trials, a computationally intensive baseline consuming \textbf{30 hours} per dataset.
In stark contrast, FlowPipe converges within \textbf{seconds} yet consistently matches or exceeds this strong baseline.
Empirically, FlowPipe achieves optimal or superior configurations in \textbf{62 out of 74} evaluated cases (spanning \textbf{14/18} on DiffPrep and \textbf{48/56} on DeepLine).
This confirms that even in high-dimensional spaces where brute-force enumeration fails, our semantic guidance effectively identifies global optima with orders-of-magnitude efficiency gains.

\textbf{Setup Ranking.} To analyze dataset-level effectiveness, we ranked the test accuracy of all 13 setups (Tables~\ref{tab5.1:diff_avg} and \ref{tab5.1:deepline_avg}). FlowPipe significantly outperforms the second-best methods, achieving average ranks of 1.139 on the DiffPrep collection (vs. CtxPipe's 3.655) and 1.089 on DeepLine (vs. SAGA's 4.132). Furthermore, despite sharing similar component candidates with baselines, FlowPipe secures the \#1 rank on approximately 78\% of DiffPrep and over 85\% of DeepLine datasets.



\textbf{Other Metrics.}
Beyond simple accuracy, we evaluate macro-averaged precision, recall, and F1-scores to assess feature quality under class imbalance. FlowPipe achieves a superior macro F1-score of \textbf{0.780} on DiffPrep and \textbf{0.805} on DeepLine, significantly outperforming the strongest baseline, CtxPipe (0.745 and 0.775, respectively). These concurrent gains confirm that FlowPipe avoids majority-class overfitting; rather, it constructs highly discriminative pipelines that effectively resolve complex decision boundaries and minimize false positives better than prior methods.


\begin{table}[h]
\centering
\caption{\reva{Performance across different downstream models.}}
\vspace{-2mm}
\label{tab:multi_model}
\resizebox{0.9\columnwidth}{!}{
\begin{tabular}{clcc}
\toprule
\textbf{Benchmark} & \textbf{Downstream Model} & \textbf{CtxPipe} & \textbf{FlowPipe} \\
\midrule
\multirow{2}{*}{DiffPrep} 
& Random Forest (RF) & 0.814 & \textbf{0.921} \\
& MLP                & 0.778 & \textbf{0.887} \\
\midrule
\multirow{2}{*}{DeepLine} 
& Random Forest (RF) & 0.829 & \textbf{0.871} \\
& MLP                & 0.826 & \textbf{0.933} \\
\bottomrule
\end{tabular}
}
\vspace{-1mm}
\end{table}


\reva{
\textbf{Robustness Across Downstream Models.} 
To prevent overfitting to the Logistic Regression evaluator, FlowPipe inherently optimizes across an ensemble of diverse predictors (\eg Random Forest, MLP, Ridge, SGD). Table~\ref{tab:multi_model} verifies this robust generalization to non-linear architectures. On DiffPrep, FlowPipe achieves average accuracies of 0.921 (Random Forest) and 0.887 (MLP), substantially outperforming CtxPipe (0.814 and 0.778, respectively). 
Similar dominance on DeepLine confirms that FlowPipe captures intrinsic data semantics to construct universally robust pipelines for a diverse spectrum of downstream models.
}

\vspace{-2mm}
\revm{
\subsubsection{Case Study}
\label{sec:case}

To elucidate FlowPipe's substantial performance margin over CtxPipe (\textbf{0.819} vs. 0.590), we qualitatively dissect pipelines generated for the \textit{google} dataset. This dataset represents a complex feature space characterized by severe power-law distributions (\eg ``Reviews'') and high-cardinality categoricals. The constructed pipelines reveal fundamental reasoning differences:

\noindent \textbf{FlowPipe (Ours)}: \texttt{Mean} $\rightarrow$ \texttt{Quantile\-Transformer} $\rightarrow$ \texttt{Interaction\-Features} \\
\noindent \textbf{CtxPipe (SOTA)}: \texttt{Mean} $\rightarrow$ \texttt{Label\-Encoder} $\rightarrow$ \texttt{Variance\-Threshold} $\rightarrow$ \texttt{Power\-Transformer} $\rightarrow$ \texttt{Random\-Trees\-Embedding}

While both identify rudimentary data hygiene needs (\eg \texttt{Mean}), subsequent selections diverge sharply. CtxPipe mechanically applies high-variance embeddings (\eg \texttt{Random-Trees-Embedding}) without resolving underlying skewness. Conversely, FlowPipe’s FiLM mechanism leverages LLM priors to recognize power-law semantics in features like ``Reviews''. It proactively prioritizes a \texttt{Quantile-Transformer}, effectively neutralizing long-tail effects via targeted semantic alignment. Beyond isolated operators, this divergence exposes flaws in sequential reasoning. CtxPipe’s decoupled Multi-DQN struggles with long-horizon credit assignment, inevitably trapping the search in sub-optimal deterministic paths. FlowPipe’s flow-based trajectory optimization overcomes this holistically, enforcing a critical prerequisite: stabilizing the feature distribution \textit{before} applying \texttt{Interaction-Features}. This safely captures non-additive relationships without amplifying noise. Ultimately, FlowPipe transcends pure combinatorial search, performing \textit{semantic navigation} tightly aligned with intrinsic data semantics.
}
\vspace{-4mm}

\begin{figure}[t]
    \centering
    \includegraphics[width=0.92\linewidth]{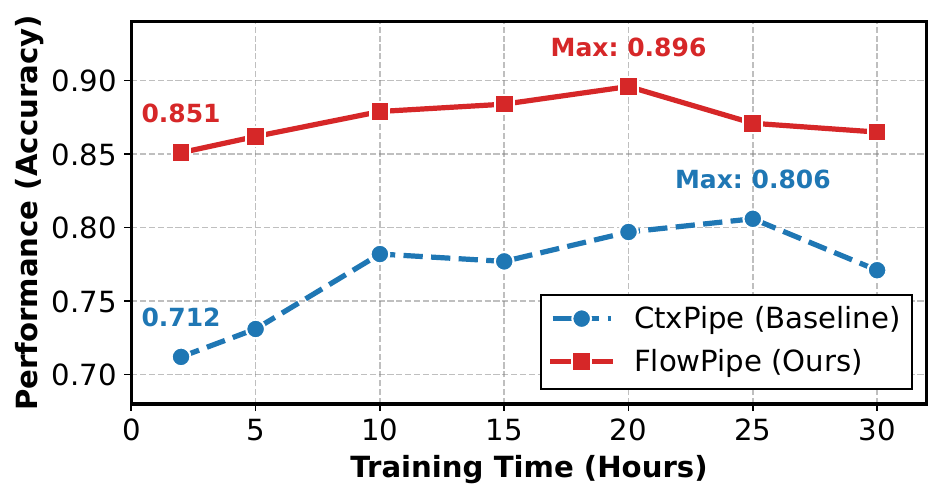}%
    \vspace{-4 mm}
    \caption{\revb{Training efficiency comparison.}}
    \vspace{-2 mm}
    \label{fig:efficiency}
\end{figure}

\subsection{Efficiency Analysis (RQ2)}
\label{sec:efficiency} 
Efficiency is paramount in production environments where continuous data streams require expedient processing~\cite{xin2021production}. FlowPipe addresses this by employing \textbf{amortized optimization}. Unlike evolutionary methods (\eg SAGA) that incur high search costs for every new dataset, FlowPipe treats training as a one-time offline investment, enabling rapid online inference.

\textbf{Training Efficiency and Amortization.}
Figure~\ref{fig:efficiency} illustrates the training efficiency on the DiffPrep collection. FlowPipe exhibits an extremely steep learning curve: it surpasses the peak performance of the strongest baseline, CtxPipe (which requires 25 hours), in just \textbf{2 hours}—a \textbf{12.5$\times$} speedup. As training saturates at approximately 20 hours, FlowPipe extends its lead to a peak accuracy of \textbf{0.896}.

\textbf{Online Inference Latency.}
We define \textit{inference time} as the wall-clock latency to synthesize a pipeline for an unseen dataset. \revc{Computationally, FlowPipe's generative state transitions scale linearly $\mathcal{O}(L)$ with pipeline depth $L$. Crucially, our \textit{decoupled prompt design} extracts strictly column metadata, rendering the generation complexity independent of dataset volume ($\mathcal{O}(1)$ w.r.t. row count $N$). Scaling to millions of rows introduces zero additional tokens. Across dimensions up to \textit{home\_credit} ($D=343$), our largest prompt consumes only $\sim$25\% of a 128K context window, ensuring robust scalability.}
\reva{To ensure rigorous evaluation, reported inference times \textit{strictly include} the execution overhead of the Llama-3.1-8B model. Empirical profiling reveals this one-time offline extraction averages just 0.383s (1.86\% of total latency)}. 
\revc{Furthermore, computing and caching these LLM priors offline grants FlowPipe a lightweight memory footprint of $\sim$1.9 GB VRAM, vastly outperforming Multi-DQN baselines (\eg CtxPipe at $\sim$4.5 GB) that require resident models during the active RL loop.}


\begin{figure}[t]
  \centering
  \includegraphics[width=0.95\linewidth]{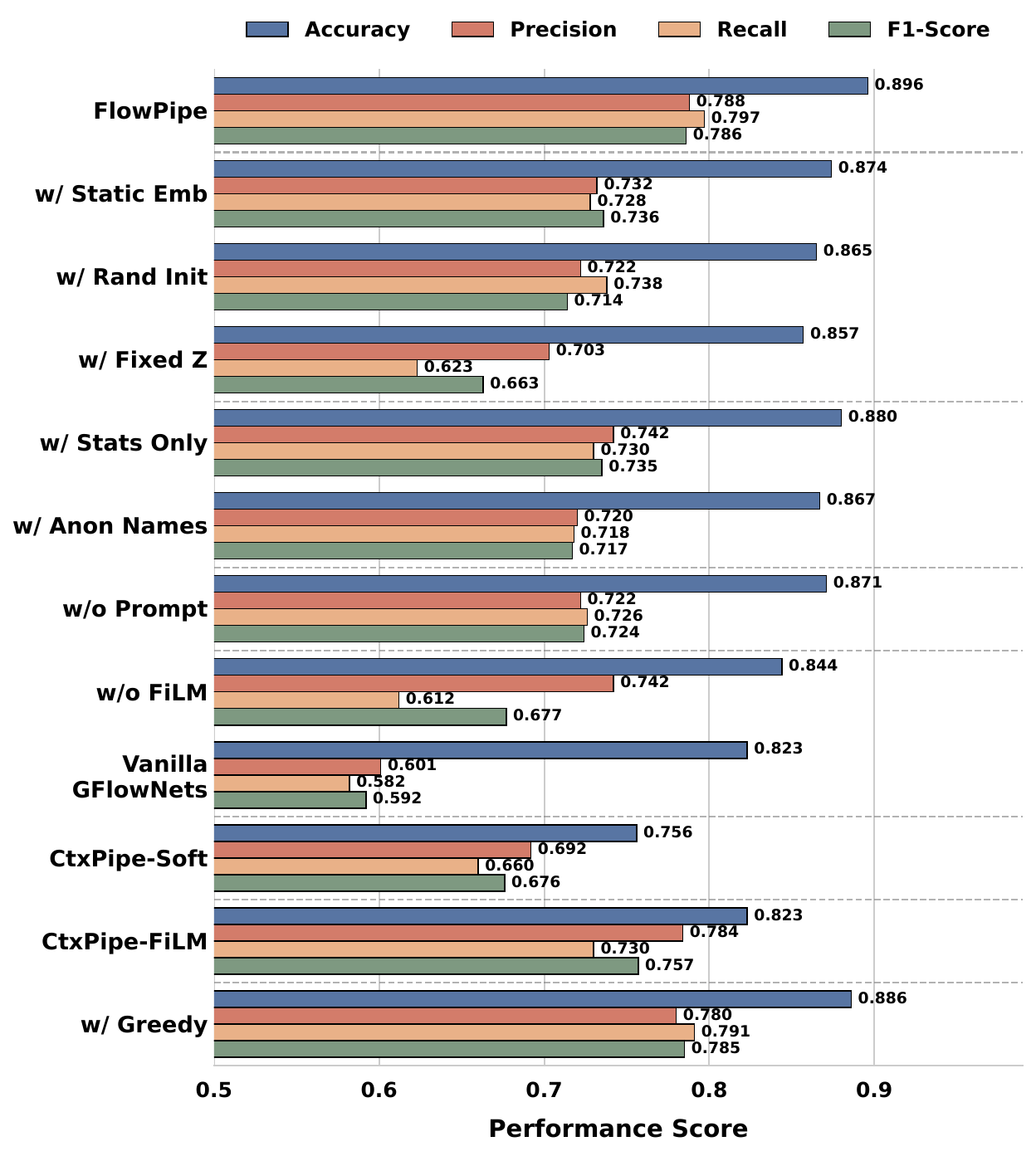}
  \vspace{-4mm}
\caption{\revc{Component ablation and baseline exploration.}}  \label{fig:ablation}
  \vspace{-5mm}
\end{figure}
On the DiffPrep collection, FlowPipe achieves an average latency of 51.516s, representing an order-of-magnitude acceleration over ``search-at-inference'' baselines like RS (803.9s) and SAGA (1384.8s). Similarly, on DeepLine, FlowPipe (10.51s) drastically outperforms the gradient-based DP-Flex ($>$2100s). While SwiftDP achieves the lowest absolute latency (5.76s) via static retrieval, this comes at a severe cost to quality (0.780 accuracy vs. 0.912 for FlowPipe), as its generalization is strictly bounded by repository coverage.

To conclude, FlowPipe establishes a win-win dominance over the strongest \ac{SOTA} baseline CtxPipe. It simultaneously outperforms CtxPipe in effectiveness (surpassing average accuracy on both benchmarks) and efficiency. Specifically, FlowPipe reduces the inference latency from 65.203s (CtxPipe) to 51.516s on DiffPrep and from 14.310s to 10.51s on DeepLine, achieving substantial speedups of \textbf{20.99\%} and \textbf{26.55\%}, respectively. This confirms FlowPipe as the new Pareto-optimal solution for automated data preparation.

\subsection{Ablation Study (RQ3)}
\label{sec:ablation}

To rigorously validate the contribution of each component in FlowPipe, we conduct a comprehensive ablation study on the DiffPrep collection. We isolate specific design choices through component replacements (\texttt{w/}) and removals (\texttt{w/o}), with comparative results visualized in Figure~\ref{fig:ablation}.


\textbf{Effectiveness of Semantic Reasoning.} Replacing the LLM reasoning module with static \textit{GTE-Large} embeddings (\texttt{w/ Static Emb}~\cite{ctxpipe}) degrades accuracy to 0.874. Similarly, using raw snapshots (\texttt{w/o Prompt}, 0.871) or random contexts (\texttt{w/ Rand Init}, 0.865) yields comparable drops. This shows passive representations fail to deduce actionable rules. Thus, these ablations confirm active LLM-derived priors are strictly essential to guarantee zero-shot generalization and prevent overfitting.

\revc{\textbf{Beyond Statistical Compression.}
To verify that the LLM performs genuine reasoning rather than mere data compression, we test two prompt variants. Restricting the prompt to pure statistics (\texttt{w/ Stats Only}) degrades accuracy to 0.880. 
Crucially, retaining the full prompt structure but obfuscating explicit column names (\texttt{w/ Anon Names}) drops accuracy further to 0.867. This proves that statistical profiles alone cannot resolve operational ambiguities; the LLM actively relies on real-world semantic anchors to map statistics into valid transformation logic.}

\textbf{Impact of Deep Modulation Mechanism.}
We evaluate the structural integration of semantic priors. Reverting the \ac{FiLM} mechanism to the shallow additive fusion used in prior works (\texttt{w/o FiLM}) results in the sharpest accuracy drop among semantic variants (0.844). This empirically validates that additive fusion is structurally inadequate for modeling complex conditional dependencies. \ac{FiLM} functions as a differentiable logical gate, explicitly magnifying or suppressing feature channels based on context; its removal causes a severe loss of policy expressiveness.

\textbf{Necessity of Dynamic Exploration Calibration.}
Replacing the learnable partition function $log \mathbf{Z}$ with a fixed scalar (\texttt{w/ Fixed Z}) degrades accuracy to 0.857 and, more critically, collapses the F1-Score to 0.663. This discrepancy validates our theoretical assertion regarding \textit{Global Flow Calibration}. Without a dynamically calibrated exploration budget, the agent converges to a conservative equilibrium, prioritizing simple, safe pipelines while failing to explore the complex, high-reward architectures necessary for optimizing F1.

\textbf{Structural Supremacy of Unified Flow Matching.}
The semantic-free \texttt{Vanilla \ac{GFlowNets}} (\textbf{0.823}) outperforms prior RL baselines, including monolithic DQN (DeepLine: \textbf{0.704}) and Multi-DQN (CtxPipe: \textbf{0.806}). This isolates the structural advantage of our framework: unlike fragmented value-based methods that struggle with delayed rewards, trajectory-level flow-matching natively propagates terminal supervision back to early decisions, guaranteeing stable and efficient long-horizon credit assignment.
\revc{\textbf{Insufficiency of Value-Based Augmentations.}
To verify our gains stem from a fundamental paradigm shift, we evaluate two targeted augmentations to the Multi-DQN baseline (Figure~\ref{fig:ablation}). First, appending FiLM for semantic integration (\texttt{CtxPipe-FiLM}) improves vanilla CtxPipe (0.823 vs.\ 0.806) but significantly trails FlowPipe (0.896), proving modulation alone cannot fix DQN's inherent long-horizon credit assignment failures. Second, an entropy-regularized variant (\texttt{CtxPipe-Soft}) intended to mimic flow-matching paradoxically \textit{degrades} accuracy (0.756 vs.\ 0.806), as forced localized stochasticity in a fragile operator space frequently triggers fatally compounding invalid actions.}

\revc{\textbf{Necessity of Flow-Based Mode Coverage.}
Disabling reward-proportional sampling in favor of deterministic decoding (\texttt{w/ Greedy}) drops the average accuracy on DiffPrep from 0.896 to 0.886. This confirms that structural mode coverage is essential to bypass local optima and discover the high-reward regions. 
For instance, on \textit{page-blocks}, FlowPipe robustly explores distinct yet equally optimal paths (\texttt{InteractionFeatures} + \texttt{Normalizer} vs. \texttt{PCA\_AUTO} + \texttt{StandardScaler}) before convergence. The top candidates evaluated during this search exhibit a minimal accuracy variance of 0.0012, proving FlowPipe reliably maps high-reward plateaus without collapsing into sub-optimal deterministic traps.}

\begin{figure}[t]
  \centering
\includegraphics[width=0.88\linewidth]{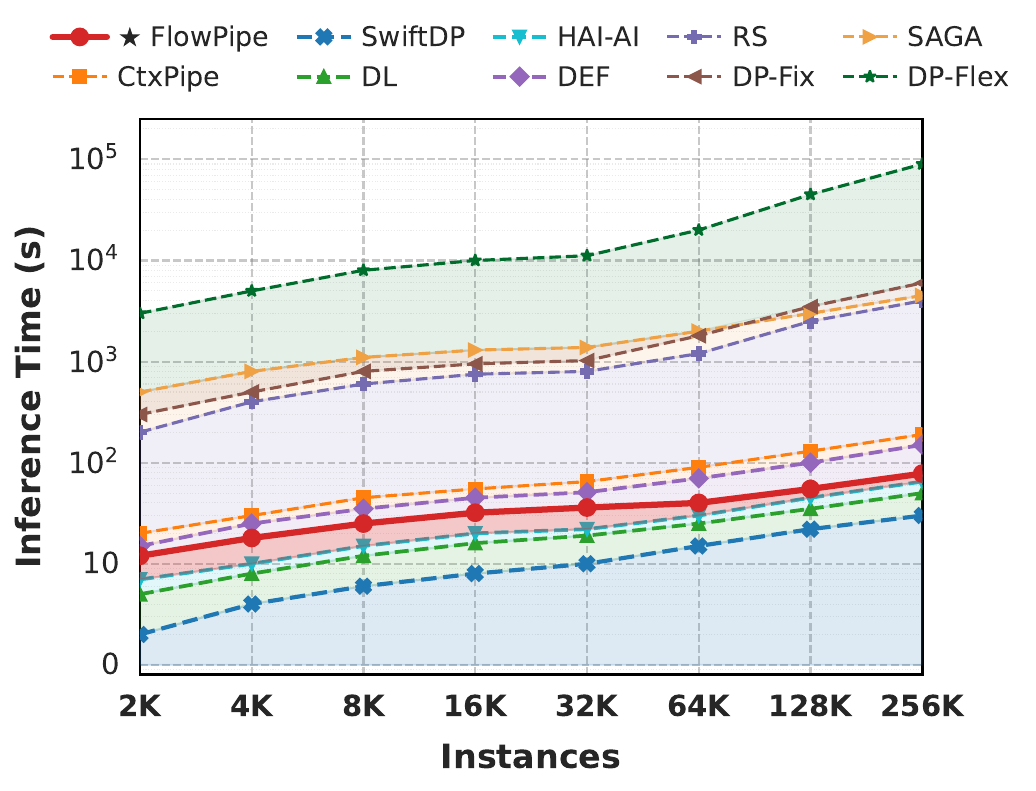}
  \vspace{-4mm}
  \caption{Scalability under different dataset sizes.}
  \label{fig:scal}
  \vspace{-3mm}
\end{figure}

\vspace{-4mm}
\subsection{Scalability and Robustness (RQ4)}
\label{sec.scal}
Finally, we address \textbf{RQ4} by examining the system's scalability under varying data loads and its robustness to hyperparameter variations.

\textbf{Row Scalability.}
We evaluate execution efficiency across 9 validated OpenML datasets, with sizes increasing exponentially from 2K to 256K instances. As plotted on a log-log scale in Figure~\ref{fig:scal}, baselines exhibit distinct performance degradation. Specifically, the gradient-based DP-Flex suffers severe computational overhead, showing an order-of-magnitude runtime spike for datasets exceeding 32K rows due to iterative backpropagation. Similarly, RS and SAGA display linear or super-linear growth driven by repetitive search. In stark contrast, FlowPipe maintains a low and stable inference time despite exponential data growth. This confirms that our amortized inference successfully decouples pipeline generation complexity from dataset volume, ensuring viability for large-scale scenarios.

\begin{figure}[t]
  \centering
  \includegraphics[width=0.95\linewidth]{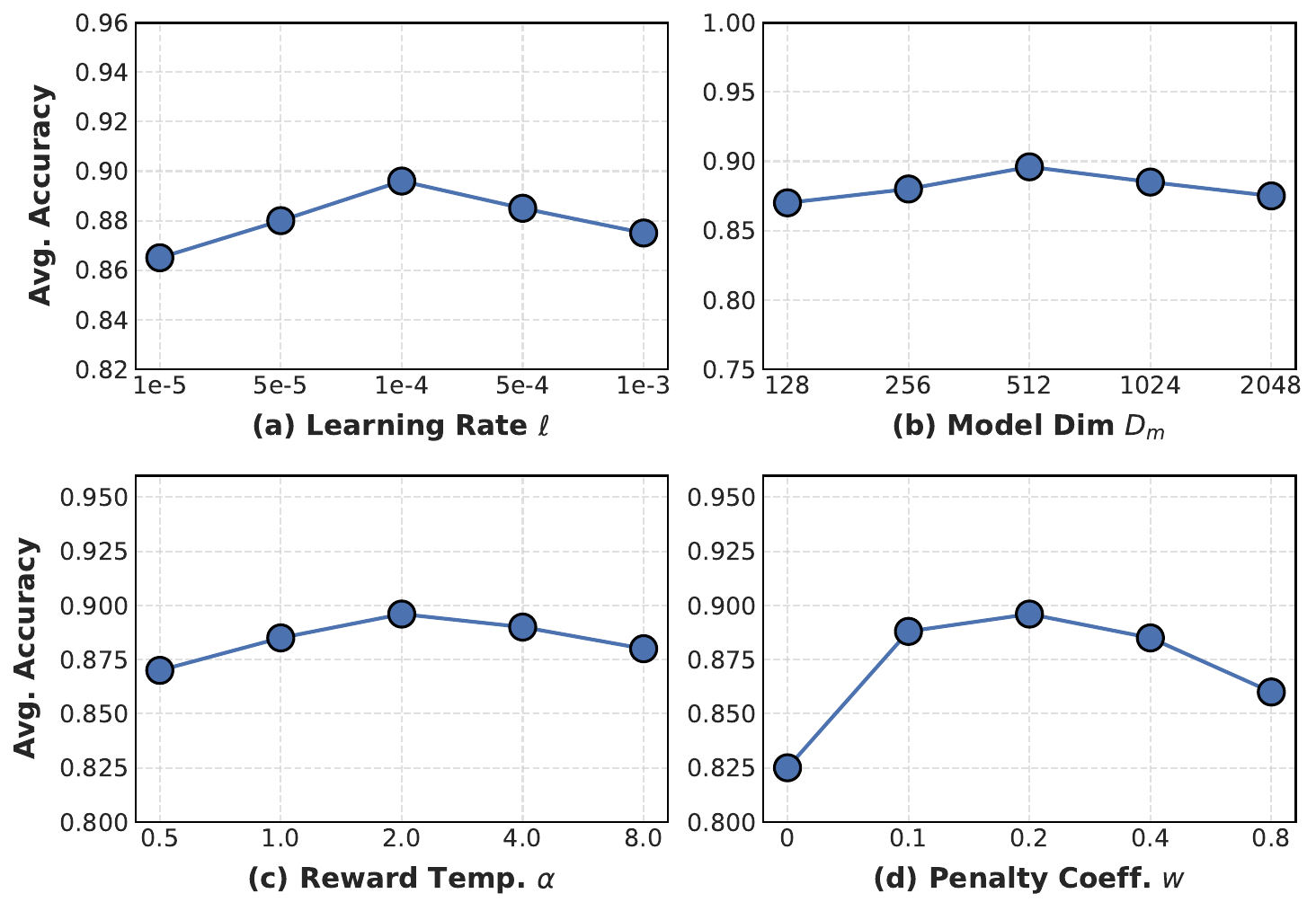}
  \caption{Hyperparameter sensitivity analysis.}
  \label{fig:hyper}
\end{figure}

\revb{\textbf{Feature Scalability.}
To isolate the impact of feature dimensionality ($D$), we compare datasets with $\sim$7K rows but varying $D$, from \texttt{HTRU\_2} ($D=9$) to \texttt{home\_credit} ($D=343$).
FlowPipe remains highly robust across this extreme range (\num{0.988} vs.\ \num{0.964} accuracy), significantly outperforming the top baseline, CtxPipe (\num{0.951} vs.\ \num{0.914}). This stability is achieved because mean pooling (Sec.~5.2) rigorously converts variable-length LLM metadata into a fixed-size conditioning vector. Consequently, structural search complexity is natively decoupled from feature width, effectively averting the curse of dimensionality.}

\begin{table}[htbp]
    \vspace{-1 mm}
    \centering
    \setlength{\tabcolsep}{4pt} 
    \caption{\revb{Scalability of Component Expansion}}
    \vspace{-2 mm}
    \label{tab:component_expansion}
    \begin{tabular}{l cccc c}
    \toprule
    \multirow{2}{*}{\textbf{Space Config.}} & \multicolumn{4}{c}{\textbf{Performance}} & \multirow{2}{*}{\textbf{Overhead}} \\
    \cmidrule(lr){2-5}
    & \textbf{Acc} & \textbf{Prec.} & \textbf{Rec.} & \textbf{F1} & \\
    \midrule
    FlowPipe & 0.896 & 0.788 & 0.794 & 0.780 & - \\
    + Category  & 0.899 & 0.794 & 0.812 & 0.803 & +4.712s \\
    + Operators & 0.898 & 0.786 & 0.794 & 0.790 & +0.009s \\
    \bottomrule
    \end{tabular}
    \vspace{-2mm}
\end{table}

\revb{\textbf{Component Extensibility.} 
As shown in Table~\ref{tab:component_expansion}, we evaluate FlowPipe's \textbf{extensibility and search space scalability} by expanding the search space with an \textit{Outlier Handling} category and advanced operators (\eg \texttt{TargetEncoder}).
FlowPipe maintains robust performance with minimal latency increases. This scalability stems from our hierarchical policy and FiLM-based gating, which reduce search complexity to $\mathcal{O}(C+K)$ and prune semantically incompatible branches. By leveraging LLM priors to bypass redundant steps, FlowPipe prevents combinatorial explosion regardless of the component pool size.}
    
\textbf{Hyperparameter Sensitivity.}
We analyze the impact of key training dynamics: learning rate $\ell$, model dimension $D_m$, reward temperature $\alpha$, and failure penalty $w$. As shown in Figure~\ref{fig:hyper}, FlowPipe demonstrates robustness across wide ranges, achieving peak performance at $\ell = 1\mathrm{e}{-4}$, $D_{m} = 512$, $\alpha = 2.0$, and $w = 0.2$. Consequently, minimal tuning is required. Notably, performance shows a stronger positive correlation with $D_m$. This indicates that a larger model dimension is critical for expanding the capacity of the semantic context embedding, enabling the agent to better assimilate complex reasoning priors from the LLM.



\vspace{-1mm}
\section{Conclusion}
\label{sec.conclude}
This paper presents FlowPipe, bridging generative flow modeling with semantic reasoning to redefine automated data preparation. By synergizing \ac{C-GFlowNets}, \ac{FiLM}-based modulation, and failure-aware exploration, it constrains combinatorial search to semantically plausible regions, resolving the inefficiencies of prior RL systems. Across 74 datasets, FlowPipe significantly outperforms \ac{SOTA} baselines, achieving an \textbf{11.96\%} accuracy gain and \textbf{12.5$\times$} faster training. 
\revc{Future work will explore \textit{dynamic semantic reconditioning} to adaptively capture evolving feature distributions.}

\clearpage
\bibliographystyle{ACM-Reference-Format}
\bibliography{SIGMOD2027}

@article{hameed2020data,
  title={Data preparation: A survey of commercial tools},
  author={Hameed, Mazhar and Naumann, Felix},
  journal={ACM sigmod record},
  volume={49},
  number={3},
  pages={18--29},
  year={2020},
  publisher={ACM New York, NY, USA}
}

@article{mozzillo2023evaluation,
  title={Evaluation of dataframe libraries for data preparation on a single machine},
  author={Mozzillo, Angelo and Zecchini, Luca and Gagliardelli, Luca and Aslam, Adeel and Bergamaschi, Sonia and Simonini, Giovanni},
  journal={arXiv preprint arXiv:2312.11122},
  year={2023}
}

@article{diffprep,
  title={Diffprep: Differentiable data preprocessing pipeline search for learning over tabular data},
  author={Li, Peng and Chen, Zhiyi and Chu, Xu and Rong, Kexin},
  journal={Proceedings of the ACM on Management of Data},
  volume={1},
  number={2},
  pages={1--26},
  year={2023},
  publisher={ACM New York, NY, USA}
}

@inproceedings{deepline,
  title={Deepline: Automl tool for pipelines generation using deep reinforcement learning and hierarchical actions filtering},
  author={Heffetz, Yuval and Vainshtein, Roman and Katz, Gilad and Rokach, Lior},
  booktitle={Proceedings of the 26th ACM SIGKDD international conference on knowledge discovery \& data mining},
  pages={2103--2113},
  year={2020}
}

@article{haipipe,
  title={Haipipe: Combining human-generated and machine-generated pipelines for data preparation},
  author={Chen, Sibei and Tang, Nan and Fan, Ju and Yan, Xuemi and Chai, Chengliang and Li, Guoliang and Du, Xiaoyong},
  journal={Proceedings of the ACM on Management of Data},
  volume={1},
  number={1},
  pages={1--26},
  year={2023},
  publisher={ACM New York, NY, USA}
}

@article{boehmsystemds,
  title={SystemDS: A declarative machine learning system for the end-to-end data science lifecycle},
  author={Boehm, Matthias and Antonov, Iulian and Baunsgaard, Sebastian and Dokter, Mark and Ginth{\"o}r, Robert and Innerebner, Kevin and Klezin, Florijan and Lindstaedt, Stefanie and Phani, Arnab and Rath, Benjamin and others},
  journal={arXiv preprint arXiv:1909.02976},
  year={2019}
}

@article{lee2017big,
  title={Big healthcare data analytics: Challenges and applications},
  author={Lee, Chonho and Luo, Zhaojing and Ngiam, Kee Yuan and Zhang, Meihui and Zheng, Kaiping and Chen, Gang and Ooi, Beng Chin and Yip, Wei Luen James},
  journal={Handbook of large-scale distributed computing in smart healthcare},
  pages={11--41},
  year={2017},
  publisher={Springer}
}

@inproceedings{luo2021mlcask,
  title={MLCask: Efficient management of component evolution in collaborative data analytics pipelines},
  author={Luo, Zhaojing and Yeung, Sai Ho and Zhang, Meihui and Zheng, Kaiping and Zhu, Lei and Chen, Gang and Fan, Feiyi and Lin, Qian and Ngiam, Kee Yuan and Ooi, Beng Chin},
  booktitle={2021 IEEE 37th International Conference on Data Engineering (ICDE)},
  pages={1655--1666},
  year={2021},
  organization={IEEE}
}

@article{saga,
  title={SAGA: A scalable framework for optimizing data cleaning pipelines for machine learning applications},
  author={Siddiqi, Shafaq and Kern, Roman and Boehm, Matthias},
  journal={Proceedings of the ACM on Management of Data},
  volume={1},
  number={3},
  pages={1--26},
  year={2023},
  publisher={ACM New York, NY, USA}
}

@inproceedings{swiftdp,
  title={SwiftDP: An Efficient Framework for Automated Data Preparation Pipeline Generation},
  author={Li, Liangwei and Zhang, Yiyi and Wang, Ning},
  booktitle={2025 IEEE 41st International Conference on Data Engineering (ICDE)},
  pages={4572--4575},
  year={2025},
  organization={IEEE}
}

@article{ctxpipe,
  title={CtxPipe: Context-aware Data Preparation Pipeline Construction for Machine Learning},
  author={Gao, Haotian and Cai, Shaofeng and Dinh, Tien Tuan Anh and Huang, Zhiyong and Ooi, Beng Chin},
  journal={Proceedings of the ACM on Management of Data},
  volume={2},
  number={6},
  pages={1--27},
  year={2024},
  publisher={ACM New York, NY, USA}
}

@article{grinsztajn2022tree,
  title={Why do tree-based models still outperform deep learning on typical tabular data?},
  author={Grinsztajn, L{\'e}o and Oyallon, Edouard and Varoquaux, Ga{\"e}l},
  journal={Advances in neural information processing systems},
  volume={35},
  pages={507--520},
  year={2022}
}

@article{tpe,
  title={Tree-structured parzen estimator: Understanding its algorithm components and their roles for better empirical performance},
  author={Watanabe, Shuhei},
  journal={arXiv preprint arXiv:2304.11127},
  year={2023}
}

@article{shwartz2022tabular,
  title={Tabular data: Deep learning is not all you need},
  author={Shwartz-Ziv, Ravid and Armon, Amitai},
  journal={Information Fusion},
  volume={81},
  pages={84--90},
  year={2022},
  publisher={Elsevier}
}

@article{abedjan2016detecting,
  title={Detecting Data Errors: Where are we and what needs to be done?},
  author={Abedjan, Ziawasch and Chul, Xu and Deng, Dong and Fernandez, Raul Castro and Ilyasl, Ihab F and Ouzzani, Mourad and Papotti, Paolo and Stonebraker, Michael and Tang, Nan},
  journal={Proceedings of the VLDB Endowment},
  volume={9},
  number={12},
  year={2016}
}

@article{zaidi2017market,
  title={Market Guide for Data Preparation},
  author={Zaidi, Ehtisham and Sallam, Rita and Vashisth, Shubhangi},
  journal={cit. on},
  pages={2},
  year={2017}
}

@article{feurer2015efficient,
  title={Efficient and robust automated machine learning},
  author={Feurer, Matthias and Klein, Aaron and Eggensperger, Katharina and Springenberg, Jost and Blum, Manuel and Hutter, Frank},
  journal={Advances in neural information processing systems},
  volume={28},
  year={2015}
}

@inproceedings{learn2clean,
  title={Learn2clean: Optimizing the sequence of tasks for web data preparation},
  author={Berti-Equille, Laure},
  booktitle={The world wide web conference},
  pages={2580--2586},
  year={2019}
}

@inproceedings{autoweka,
  title={Auto-WEKA: Combined selection and hyperparameter optimization of classification algorithms},
  author={Thornton, Chris and Hutter, Frank and Hoos, Holger H and Leyton-Brown, Kevin},
  booktitle={Proceedings of the 19th ACM SIGKDD international conference on Knowledge discovery and data mining},
  pages={847--855},
  year={2013}
}

@inproceedings{TPOT,
  title={TPOT: A tree-based pipeline optimization tool for automating machine learning},
  author={Olson, Randal S and Moore, Jason H},
  booktitle={Workshop on automatic machine learning},
  pages={66--74},
  year={2016},
  organization={PMLR}
}

@inproceedings{chai,
  title={Demystifying artificial intelligence for data preparation},
  author={Chai, Chengliang and Tang, Nan and Fan, Ju and Luo, Yuyu},
  booktitle={Companion of the 2023 International Conference on Management of Data},
  pages={13--20},
  year={2023}
}

@article{bengio2021flow,
  title={Flow network based generative models for non-iterative diverse candidate generation},
  author={Bengio, Emmanuel and Jain, Moksh and Korablyov, Maksym and Precup, Doina and Bengio, Yoshua},
  journal={Advances in neural information processing systems},
  volume={34},
  pages={27381--27394},
  year={2021}
}

@article{bengio2023gflownet,
  title={Gflownet foundations},
  author={Bengio, Yoshua and Lahlou, Salem and Deleu, Tristan and Hu, Edward J and Tiwari, Mo and Bengio, Emmanuel},
  journal={Journal of Machine Learning Research},
  volume={24},
  number={210},
  pages={1--55},
  year={2023}
}

@article{kumar2020conservative,
  title={Conservative q-learning for offline reinforcement learning},
  author={Kumar, Aviral and Zhou, Aurick and Tucker, George and Levine, Sergey},
  journal={Advances in neural information processing systems},
  volume={33},
  pages={1179--1191},
  year={2020}
}

@article{zhang2023let,
  title={Let the flows tell: Solving graph combinatorial problems with gflownets},
  author={Zhang, Dinghuai and Dai, Hanjun and Malkin, Nikolay and Courville, Aaron C and Bengio, Yoshua and Pan, Ling},
  journal={Advances in neural information processing systems},
  volume={36},
  pages={11952--11969},
  year={2023}
}

@inproceedings{jain2022biological,
  title={Biological sequence design with gflownets},
  author={Jain, Moksh and Bengio, Emmanuel and Hernandez-Garcia, Alex and Rector-Brooks, Jarrid and Dossou, Bonaventure FP and Ekbote, Chanakya Ajit and Fu, Jie and Zhang, Tianyu and Kilgour, Michael and Zhang, Dinghuai and others},
  booktitle={International Conference on Machine Learning},
  pages={9786--9801},
  year={2022},
  organization={PMLR}
}

@article{narayan2022can,
  title={Can Foundation Models Wrangle Your Data?},
  author={Narayan, Avanika and Chami, Ines and Orr, Laurel and R{\'e}, Christopher},
  journal={Proceedings of the VLDB Endowment},
  volume={16},
  number={4},
  pages={738--746},
  year={2022},
  publisher={VLDB Endowment}
}

@article{zhang2023large,
  title={Large language models as data preprocessors},
  author={Zhang, Haochen and Dong, Yuyang and Xiao, Chuan and Oyamada, Masafumi},
  journal={arXiv preprint arXiv:2308.16361},
  year={2023}
}

@article{li2024table,
  title={Table-gpt: Table fine-tuned gpt for diverse table tasks},
  author={Li, Peng and He, Yeye and Yashar, Dror and Cui, Weiwei and Ge, Song and Zhang, Haidong and Rifinski Fainman, Danielle and Zhang, Dongmei and Chaudhuri, Surajit},
  journal={Proceedings of the ACM on Management of Data},
  volume={2},
  number={3},
  pages={1--28},
  year={2024},
  publisher={ACM New York, NY, USA}
}

@inproceedings{chen2024chatpipe,
  title={Chatpipe: Orchestrating data preparation pipelines by optimizing human-chatgpt interactions},
  author={Chen, Sibei and Liu, Hanbing and Jin, Waiting and Sun, Xiangyu and Feng, Xiaoyao and Fan, Ju and Du, Xiaoyong and Tang, Nan},
  booktitle={Companion of the 2024 International Conference on Management of Data},
  pages={484--487},
  year={2024}
}

@inproceedings{yin2020tabert,
  title={TaBERT: Pretraining for Joint Understanding of Textual and Tabular Data},
  author={Yin, Pengcheng and Neubig, Graham and Yih, Wen-tau and Riedel, Sebastian},
  booktitle={Proceedings of the 58th Annual Meeting of the Association for Computational Linguistics},
  pages={8413--8426},
  year={2020}
}

@inproceedings{chu2016data,
  title={Data cleaning: Overview and emerging challenges},
  author={Chu, Xu and Ilyas, Ihab F and Krishnan, Sanjay and Wang, Jiannan},
  booktitle={Proceedings of the 2016 international conference on management of data},
  pages={2201--2206},
  year={2016}
}

@inproceedings{xin2021production,
  title={Production machine learning pipelines: Empirical analysis and optimization opportunities},
  author={Xin, Doris and Miao, Hui and Parameswaran, Aditya and Polyzotis, Neoklis},
  booktitle={Proceedings of the 2021 international conference on management of data},
  pages={2639--2652},
  year={2021}
}

@inproceedings{shang2019democratizing,
  title={Democratizing data science through interactive curation of ml pipelines},
  author={Shang, Zeyuan and Zgraggen, Emanuel and Buratti, Benedetto and Kossmann, Ferdinand and Eichmann, Philipp and Chung, Yeounoh and Binnig, Carsten and Upfal, Eli and Kraska, Tim},
  booktitle={Proceedings of the 2019 international conference on management of data},
  pages={1171--1188},
  year={2019}
}

@article{vanschoren2014openml,
  title={OpenML: networked science in machine learning},
  author={Vanschoren, Joaquin and Van Rijn, Jan N and Bischl, Bernd and Torgo, Luis},
  journal={ACM SIGKDD Explorations Newsletter},
  volume={15},
  number={2},
  pages={49--60},
  year={2014},
  publisher={ACM New York, NY, USA}
}

@inproceedings{van2016deep,
  title={Deep reinforcement learning with double q-learning},
  author={Van Hasselt, Hado and Guez, Arthur and Silver, David},
  booktitle={Proceedings of the AAAI conference on artificial intelligence},
  volume={30},
  number={1},
  year={2016}
}

@misc{asuncion2007uci,
  title={UCI machine learning repository},
  author={Asuncion, Arthur and Newman, David and others},
  year={2007},
  publisher={Irvine, CA, USA}
}

@article{sculley2015hidden,
  title={Hidden technical debt in machine learning systems},
  author={Sculley, David and Holt, Gary and Golovin, Daniel and Davydov, Eugene and Phillips, Todd and Ebner, Dietmar and Chaudhary, Vinay and Young, Michael and Crespo, Jean-Francois and Dennison, Dan},
  journal={Advances in neural information processing systems},
  volume={28},
  year={2015}
}

@article{polyzotis2019data,
  title={Data validation for machine learning},
  author={Polyzotis, Neoklis and Zinkevich, Martin and Roy, Sudip and Breck, Eric and Whang, Steven},
  journal={Proceedings of machine learning and systems},
  volume={1},
  pages={334--347},
  year={2019}
}

@inproceedings{perez2018film,
  title={Film: Visual reasoning with a general conditioning layer},
  author={Perez, Ethan and Strub, Florian and De Vries, Harm and Dumoulin, Vincent and Courville, Aaron},
  booktitle={Proceedings of the AAAI conference on artificial intelligence},
  volume={32},
  number={1},
  year={2018}
}

@article{li2018hyperband,
  title={Hyperband: A novel bandit-based approach to hyperparameter optimization},
  author={Li, Lisha and Jamieson, Kevin and DeSalvo, Giulia and Rostamizadeh, Afshin and Talwalkar, Ameet},
  journal={Journal of Machine Learning Research},
  volume={18},
  number={185},
  pages={1--52},
  year={2018}
}

@inproceedings{yang2019oboe,
  title={OBOE: Collaborative filtering for AutoML model selection},
  author={Yang, Chengrun and Akimoto, Yuji and Kim, Dae Won and Udell, Madeleine},
  booktitle={Proceedings of the 25th ACM SIGKDD international conference on knowledge discovery \& data mining},
  pages={1173--1183},
  year={2019}
}

@inproceedings{liudarts,
  title={DARTS: Differentiable Architecture Search},
  author={Liu, Hanxiao and Simonyan, Karen and Yang, Yiming},
  booktitle={International Conference on Learning Representations}
}

@inproceedings{li2020differentiable,
  title={Differentiable automatic data augmentation},
  author={Li, Yonggang and Hu, Guosheng and Wang, Yongtao and Hospedales, Timothy and Robertson, Neil M and Yang, Yongxin},
  booktitle={European conference on computer vision},
  pages={580--595},
  year={2020},
  organization={Springer}
}

@inproceedings{fan2020autofs,
  title={Autofs: Automated feature selection via diversity-aware interactive reinforcement learning},
  author={Fan, Wei and Liu, Kunpeng and Liu, Hao and Wang, Pengyang and Ge, Yong and Fu, Yanjie},
  booktitle={2020 IEEE International Conference on Data Mining (ICDM)},
  pages={1008--1013},
  year={2020},
  organization={IEEE}
}

@inproceedings{khurana2018feature,
  title={Feature engineering for predictive modeling using reinforcement learning},
  author={Khurana, Udayan and Samulowitz, Horst and Turaga, Deepak},
  booktitle={Proceedings of the AAAI conference on artificial intelligence},
  volume={32},
  number={1},
  year={2018}
}

@article{yangauto,
  title={Auto-pipeline: synthesizing complex data pipelines by-target using reinforcement learning and search},
  author={Yang, Junwen and He, Yeye and Chaudhuri, Surajit},
  journal={Proceedings of the VLDB Endowment},
  volume={14},
  number={11},
  pages={2563--2575},
  year={2021},
  publisher={VLDB Endowment}
}

@article{phamefficient,
  title={Efficient Neural Architecture Search via Parameter Sharing},
  author={Pham, Hieu and Guan, Melody Y and Zoph, Barret and Le, Quoc V and Dean, Jeff}
}

@article{li2023towards,
  title={Towards general text embeddings with multi-stage contrastive learning},
  author={Li, Zehan and Zhang, Xin and Zhang, Yanzhao and Long, Dingkun and Xie, Pengjun and Zhang, Meishan},
  journal={arXiv preprint arXiv:2308.03281},
  year={2023}
}

@article{yu2021windtunnel,
  title={WindTunnel: towards differentiable ML pipelines beyond a single model},
  author={Yu, Gyeong-In and Amizadeh, Saeed and Kim, Sehoon and Pagnoni, Artidoro and Zhang, Ce and Chun, Byung-Gon and Weimer, Markus and Interlandi, Matteo},
  journal={Proceedings of the VLDB Endowment},
  volume={15},
  number={1},
  pages={11--20},
  year={2021},
  publisher={VLDB Endowment}
}

@inproceedings{hilprecht2023diffml,
  title={Diffml: End-to-end differentiable ML pipelines},
  author={Hilprecht, Benjamin and Hammacher, Christian and Reis, Eduardo S and Abdelaal, Mohamed and Binnig, Carsten},
  booktitle={Proceedings of the Seventh Workshop on Data Management for End-to-End Machine Learning},
  pages={1--7},
  year={2023}
}

@article{schulman2017proximal,
  title={Proximal policy optimization algorithms},
  author={Schulman, John and Wolski, Filip and Dhariwal, Prafulla and Radford, Alec and Klimov, Oleg},
  journal={arXiv preprint arXiv:1707.06347},
  year={2017}
}

@article{chang2025shapleypipe,
  title={ShapleyPipe: Hierarchical Shapley Search for Data Preparation Pipeline Construction},
  author={Chang, Jing and Liu, Chang and Huang, Jinbin and Zheng, Shuyuan and Mao, Rui and Qin, Jianbin},
  journal={arXiv preprint arXiv:2510.27168},
  year={2025}
}

@article{grattafiori2024llama,
  title={The llama 3 herd of models},
  author={Grattafiori, Aaron and Dubey, Abhimanyu and Jauhri, Abhinav and Pandey, Abhinav and Kadian, Abhishek and Al-Dahle, Ahmad and Letman, Aiesha and Mathur, Akhil and Schelten, Alan and Vaughan, Alex and others},
  journal={arXiv preprint arXiv:2407.21783},
  year={2024}
}

@inproceedings{silva2025gflownets,
  title={When do GFlowNets learn the right distribution?},
  author={Silva, Tiago and Alves, Rodrigo Barreto and da Silva, Eliezer de Souza and Souza, Amauri H and Garg, Vikas and Kaski, Samuel and Mesquita, Diego},
  booktitle={The Thirteenth International Conference on Learning Representations},
  year={2025}
}

@inproceedings{niu2024gflownet,
  title={GFlowNet training by policy gradients},
  author={Niu, Puhua and Wu, Shili and Fan, Mingzhou and Qian, Xiaoning},
  booktitle={Proceedings of the 41st International Conference on Machine Learning},
  pages={38344--38380},
  year={2024}
}

@inproceedings{hu2023gflownet,
  title={GFlowNet-EM for learning compositional latent variable models},
  author={Hu, Edward J and Malkin, Nikolay and Jain, Moksh and Everett, Katie E and Graikos, Alexandros and Bengio, Yoshua},
  booktitle={International Conference on Machine Learning},
  pages={13528--13549},
  year={2023},
  organization={PMLR}
}

@article{schelter2018automating,
  title={Automating large-scale data quality verification},
  author={Schelter, Sebastian and Lange, Dustin and Schmidt, Philipp and Celikel, Meltem and Biessmann, Felix and Grafberger, Andreas},
  journal={Proceedings of the VLDB Endowment},
  volume={11},
  number={12},
  pages={1781--1794},
  year={2018},
  publisher={VLDB Endowment}
}

@article{fathollahzadeh2025catdb,
  title={CatDB: Data-catalog-guided, LLM-based Generation of Data-centric ML Pipelines},
  author={Fathollahzadeh, Saeed and Mansour, Essam and Boehm, Matthias},
  journal={Proceedings of the VLDB Endowment},
  volume={18},
  number={8},
  pages={2639--2652},
  year={2025},
  publisher={VLDB Endowment}
}

@inproceedings{ledell2020h2o,
  title={H2o automl: Scalable automatic machine learning},
  author={LeDell, Erin and Poirier, Sebastien and others},
  booktitle={Proceedings of the AutoML Workshop at ICML},
  volume={2020},
  pages={24},
  year={2020}
}

@incollection{kang2024azure,
  title={Azure Automated Machine Learning},
  author={Kang, Min Soo and Park, Sung Yul and Chung, Myung-Ae and Han, Dong-hun},
  booktitle={NO-CODE AI: Concepts and Applications in Machine Learning, Visualization, and Cloud Platforms},
  pages={263--282},
  year={2024},
  publisher={World Scientific}
}

@inproceedings{bert,
  title={Bert: Pre-training of deep bidirectional transformers for language understanding},
  author={Devlin, Jacob and Chang, Ming-Wei and Lee, Kenton and Toutanova, Kristina},
  booktitle={Proceedings of the 2019 conference of the North American chapter of the association for computational linguistics: human language technologies, volume 1 (long and short papers)},
  pages={4171--4186},
  year={2019}
}
\clearpage
\newpage
\appendix
\section{Supplementary Results}
\label{sec:appendix_detailed_results}

In this appendix, we provide the full evaluation results of FlowPipe across the DiffPrep and DeepLine datasets. These results substantiate the claims made in Section~\ref{sec:exp_performance}, confirming FlowPipe's consistent superiority across a wide range of datasets. The results are presented in Tables~\ref{tab:diffprep} and \ref{tab:deepline}, where \textbf{bold} values represent the highest accuracy, while \underline{underlined} values represent the second-highest.

\subsection{Performance Overview}
Tables~\ref{tab:diffprep} and \ref{tab:deepline} summarize the individual test accuracies across the DiffPrep and DeepLine collections. In the DiffPrep dataset collection, FlowPipe achieves the highest test accuracy on \textbf{14 out of 18 datasets}, while in the DeepLine collection, FlowPipe outperforms all other methods on \textbf{48 out of 56 datasets}. This consistent performance advantage across multiple datasets reinforces the robustness of FlowPipe as a data preparation framework. 

FlowPipe’s average test accuracies are \textbf{0.896} on DiffPrep and \textbf{0.912} on DeepLine, surpassing previous \ac{SOTA} methods that achieved 0.806 and 0.813, respectively. The significant performance improvement can be attributed to FlowPipe's ability to incorporate semantic information, enhancing its ability to handle high-dimensional and complex datasets. 

\subsection{Detailed Analysis of Dataset Characteristics}
\textbf{High-Dimensional and Combinatorial Spaces}
Datasets with a large number of features, such as \textit{bureau} (302 features) and \textit{house} (81 features), often present a challenge due to the exponentially growing search space. FlowPipe demonstrates its efficiency in these high-dimensional spaces by achieving \textbf{0.965} on \textit{bureau} and \textbf{0.960} on \textit{house}. Other methods, including DP-Fix and SwiftDP, experience a decline in accuracy, indicating their difficulty in handling the increased complexity. FlowPipe, by focusing its exploration on promising regions of the space, avoids the exploration inefficiencies faced by traditional methods.

\textbf{Handling Missing Values and Outliers}
FlowPipe's ability to manage datasets with significant missing values is evident in its performance on datasets like \textit{google} and \textit{connect-4}. On \textit{google}, where 2.19\% of the data is missing, FlowPipe achieves \textbf{0.819}, outperforming the next best baseline (DP-Fix) with \textbf{0.672}. This demonstrates the effectiveness of FlowPipe's Semantic-Modulated Policy, which leverages global contextual information to guide imputation strategies. Similarly, on \textit{connect-4}, characterized by high outlier counts, FlowPipe achieves \textbf{0.891}, a clear improvement over other methods, which struggle to handle such data irregularities.

\textbf{Small Sample Sizes}
FlowPipe also excels in small-sample settings. For example, on the \textit{Iris} dataset, which contains only 150 instances, FlowPipe achieves perfect accuracy (\textbf{1.000}), a notable improvement over other methods that fail to generalize effectively in such low-data regimes. This highlights FlowPipe's ability to leverage pre-trained knowledge and semantic priors to regularize the pipeline synthesis process, avoiding overfitting on small datasets.

\subsection{Benchmarking Against Exhaustive Search ($ES^*$)}
To assess the efficacy of FlowPipe in approximating optimal pipeline configurations, we benchmark it against an Exhaustive Search ($ES^*$) baseline. $ES^*$ serves as a proxy for the empirical upper bound, utilizing a computational budget of 10,000 random evaluations per dataset. Despite the extensive search space coverage of $ES^*$, FlowPipe consistently matches or exceeds its performance. This advantage is particularly pronounced in high-dimensional tasks where the search space is vast and sparse, rendering unguided stochastic search inefficient.

\vspace{0.5em}
\noindent\textbf{Case Study: The \textit{google} Dataset.} The disparity in search efficiency is best illustrated by the \textit{google} dataset results. $ES^*$ achieves an F1-score of only \underline{0.654}, whereas FlowPipe achieves a significantly higher \textbf{0.820}. This performance gap stems from the semantic quality of the synthesized pipelines. 

As detailed in the experimental logs, the distinct configurations identified are:

\begin{itemize}[leftmargin=*]
    \item \textbf{FlowPipe}: \texttt{ImputerMean}, \texttt{Quantile\-Transformer}, \texttt{Interaction\-Features}
    \item \textbf{$ES^*$}: \texttt{ImputerMedian}, \texttt{Polynomial\-Features}, \texttt{Power\-Transformer}
\end{itemize}

The FlowPipe selection demonstrates high semantic coherence. The \textit{google} dataset contains features with heavy-tailed distributions (e.g., \texttt{Reviews}). FlowPipe correctly prioritized \texttt{Quantile\-Transformer}, which effectively mitigates outliers through non-linear mapping. The subsequent \texttt{Interaction\-Features} captures non-additive effects, maximizing information gain.

In contrast, the $ES^*$ baseline's inclusion of \texttt{Polynomial\-Features} on raw data likely introduced excessive dimensionality and noise. Generating high-degree polynomials \textit{before} stabilization (via \texttt{Power\-Transformer}) is a suboptimal sequence that exacerbates variance.

This comparison underscores that FlowPipe does not merely search; it navigates. By leveraging semantic guidance, it avoids computationally expensive regions, steering towards robust configurations within seconds.

\vspace{0.5em}
\noindent\textbf{Summary.}
The results confirm that FlowPipe is a robust solution for automated data preparation. Its performance consistently surpasses baseline methods, particularly in complex, high-dimensional regimes. By effectively integrating semantic knowledge, FlowPipe establishes itself as a state-of-the-art (\ac{SOTA}) tool for intelligent pipeline generation.

\begin{table*}[htbp]
\centering
\caption{Characteristics of the DeepLine dataset collection and comparison of test accuracy}
\vspace{1mm}

\setlength{\tabcolsep}{1.5pt}

\resizebox{\textwidth}{!}{%
    \begin{tabular}{l ccccc ccccccccccc}
    \toprule
    
     & \multicolumn{5}{c}{\textbf{Data Characteristics}} & \multicolumn{11}{c}{\textbf{Test Accuracy}} \\
    
    \cmidrule(lr){2-6} \cmidrule(lr){7-17}
    
    \textbf{Dataset} & \textbf{\#Inst.} & \textbf{\#Feat.} & \textbf{\#Lbl.} & \textbf{\%Miss.} & \textbf{\%Cat.} & \textbf{(ES$^*$)} & \textbf{DEF} & \textbf{RS} & \textbf{DP-Fix} & \textbf{DP-Flex} & \textbf{DL} & \textbf{HAI-AI} & \textbf{SAGA} & \textbf{CtxPipe} & \textbf{SwiftDP} & \textbf{Ours} \\
    \midrule
    
    Accident & 7500 & 14 & 4 & 0 & 0 & 0.665 & 0.673 & 0.674 & 0.673 & 0.674 & 0.598 & 0.600 & 0.597 & \underline{0.692} & 0.657 & \textbf{0.978} \\
    adult & 7480 & 15 & 2 & 0 & 60 & \underline{0.884} & 0.849 & 0.851 & 0.857 & 0.855 & 0.737 & 0.815 & 0.843 & 0.829 & 0.824 & \textbf{0.918} \\
    broadwaymult & 285 & 8 & 7 & 1.2 & 50 & 0.415 & 0.298 & \underline{0.561} & 0.474 & 0.526 & 0.175 & 0.456 & 0.474 & 0.512 & 0.333 & \textbf{0.749} \\
    germangss & 400 & 6 & 17 & 0 & 66.7 & 0.388 & 0.238 & 0.225 & 0.200 & 0.188 & 0.275 & 0.212 & 0.213 & \underline{0.442} & 0.275 & \textbf{0.536} \\
    ar4 & 107 & 30 & 2 & 0 & 0 & 0.895 & \underline{0.905} & 0.810 & 0.857 & 0.810 & 0.818 & 0.773 & 0.773 & 0.801 & \textbf{0.909} & \underline{0.905} \\
    bank-full & 7368 & 15 & 2 & 0 & 53.3 & \underline{0.915} & 0.886 & 0.891 & 0.896 & 0.899 & 0.840 & 0.890 & 0.896 & 0.890 & 0.886 & \textbf{0.948} \\
    baseball & 1340 & 17 & 3 & 0.1 & 5.9 & \underline{0.962} & 0.937 & 0.944 & 0.925 & 0.925 & 0.937 & 0.955 & 0.948 & 0.931 & 0.952 & \textbf{0.979} \\
    biodeg & 1055 & 42 & 2 & 0 & 2.4 & \underline{0.910} & 0.829 & 0.806 & 0.810 & 0.829 & 0.834 & 0.877 & 0.896 & 0.871 & 0.863 & \textbf{0.934} \\
    blood-transfusion & 748 & 5 & 2 & 0 & 0 & \underline{0.812} & 0.752 & 0.732 & 0.785 & 0.732 & 0.627 & 0.713 & 0.720 & 0.776 & 0.767 & \textbf{0.891} \\
    bng\_cmc & 55296 & 10 & 3 & 0 & 0 & \underline{0.654} & 0.520 & 0.535 & 0.533 & 0.537 & 0.527 & 0.525 & 0.522 & 0.648 & 0.521 & \textbf{0.878} \\
    bodyfat & 252 & 15 & 2 & 0 & 6.7 & \underline{0.992} & 0.900 & 0.900 & 0.960 & 0.940 & 0.784 & 0.980 & \textbf{1.000} & 0.941 & \textbf{1.000} & \textbf{1.000} \\
    braziltourism & 412 & 9 & 7 & 2.6 & 0 & \underline{0.815} & 0.707 & 0.683 & 0.707 & 0.695 & 0.735 & 0.771 & 0.711 & 0.745 & 0.735 & \textbf{0.885} \\
    breast & 699 & 10 & 2 & 0.2 & 10 & 0.982 & 0.971 & 0.950 & 0.964 & \underline{0.986} & 0.900 & 0.979 & 0.971 & 0.938 & 0.950 & \textbf{0.989} \\
    bureau & 7125 & 302 & 2 & 30.6 & 5.3 & \underline{0.942} & 0.928 & 0.929 & 0.911 & 0.924 & 0.923 & 0.930 & 0.922 & 0.917 & 0.915 & \textbf{0.965} \\
    car & 1728 & 7 & 4 & 0 & 100 & 0.885 & 0.852 & 0.913 & 0.916 & 0.919 & 0.963 & 0.968 & \underline{0.974} & 0.953 & 0.691 & \textbf{0.990} \\
    chatfield\_4 & 235 & 13 & 2 & 0 & 7.7 & \underline{0.936} & 0.915 & 0.915 & 0.894 & 0.872 & 0.660 & 0.894 & 0.872 & 0.859 & \textbf{0.957} & 0.923 \\
    cmc & 1473 & 10 & 3 & 0 & 0 & \underline{0.612} & 0.520 & 0.565 & 0.561 & 0.561 & 0.502 & 0.512 & 0.539 & 0.598 & 0.539 & \textbf{0.746} \\
    credit & 1000 & 21 & 2 & 0 & 66.7 & \underline{0.825} & 0.725 & 0.740 & 0.765 & 0.765 & 0.690 & 0.705 & 0.785 & 0.780 & 0.710 & \textbf{0.894} \\
    crx & 690 & 16 & 2 & 0.6 & 62.5 & 0.915 & 0.899 & 0.906 & 0.884 & \underline{0.920} & 0.870 & 0.790 & 0.877 & 0.867 & 0.812 & \textbf{0.921} \\
    banknote & 1372 & 5 & 2 & 0 & 0 & \underline{0.995} & 0.971 & 0.967 & 0.942 & 0.985 & 0.851 & \textbf{1.000} & 0.993 & 0.970 & 0.971 & \textbf{1.000} \\
    dermatology & 365 & 35 & 6 & 0.1 & 0 & \underline{0.979} & 0.959 & 0.959 & 0.945 & 0.973 & \textbf{0.986} & 0.932 & 0.973 & 0.950 & 0.973 & \textbf{0.986} \\
    diggle\_table\_a2 & 310 & 9 & 2 & 0 & 11.1 & \underline{0.984} & 0.855 & 0.952 & 0.952 & 0.935 & 0.871 & \textbf{1.000} & \textbf{1.000} & 0.970 & 0.968 & \textbf{1.000} \\
    disclosure\_z & 662 & 4 & 2 & 0 & 25 & \underline{0.645} & 0.561 & 0.492 & 0.561 & 0.598 & 0.564 & 0.496 & 0.504 & 0.616 & 0.571 & \textbf{0.771} \\
    eucalyptus & 736 & 20 & 5 & 3 & 30 & \underline{0.712} & 0.592 & 0.633 & 0.571 & 0.687 & 0.655 & 0.608 & 0.649 & 0.597 & 0.635 & \textbf{0.788} \\
    Frogs\_family & 7195 & 23 & 4 & 0 & 4.3 & \underline{0.988} & 0.938 & 0.945 & 0.944 & 0.952 & 0.905 & 0.985 & 0.985 & 0.917 & 0.947 & \textbf{0.994} \\
    Frogs\_gen & 7195 & 23 & 8 & 0 & 4.3 & \underline{0.988} & 0.945 & 0.951 & 0.950 & 0.955 & 0.927 & 0.976 & 0.984 & 0.926 & 0.953 & \textbf{0.993} \\
    Frogs\_spec & 7195 & 23 & 10 & 0 & 4.3 & \underline{0.987} & 0.954 & 0.958 & 0.958 & 0.959 & 0.943 & 0.979 & 0.984 & 0.932 & 0.969 & \textbf{0.993} \\
    glass & 214 & 10 & 6 & 0 & 0 & 0.780 & 0.571 & 0.667 & 0.571 & 0.643 & 0.605 & 0.581 & \textbf{0.837} & 0.710 & 0.744 & \underline{0.807} \\
    haberman & 306 & 4 & 2 & 0 & 0 & \underline{0.795} & 0.738 & 0.738 & 0.754 & 0.770 & 0.694 & 0.645 & 0.710 & 0.705 & 0.677 & \textbf{0.823} \\
    home\_credit & 7250 & 343 & 2 & 19.3 & 0 & \underline{0.945} & 0.928 & 0.928 & 0.897 & 0.922 & 0.917 & 0.931 & 0.922 & 0.914 & 0.907 & \textbf{0.964} \\
    HTRU\_2 & 7000 & 9 & 2 & 0 & 0 & \underline{0.984} & 0.975 & 0.973 & 0.976 & 0.979 & 0.973 & 0.974 & 0.977 & 0.951 & 0.983 & \textbf{0.988} \\
    ilpd & 583 & 11 & 2 & 0 & 9.1 & 0.695 & 0.733 & 0.741 & 0.716 & 0.716 & 0.718 & 0.709 & 0.744 & \underline{0.753} & 0.692 & \textbf{0.832} \\
    image\_seg & 2310 & 20 & 7 & 0 & 5 & 0.975 & 0.885 & 0.939 & 0.918 & 0.931 & 0.973 & \underline{0.978} & \textbf{0.981} & 0.899 & 0.944 & \textbf{0.981} \\
    Indian\_Liver & 583 & 11 & 2 & 0.1 & 0 & 0.695 & 0.741 & 0.724 & 0.707 & 0.724 & 0.598 & 0.684 & 0.752 & \underline{0.753} & 0.692 & \textbf{0.857} \\
    Iris & 150 & 5 & 3 & 0 & 20 & \underline{0.980} & 0.800 & 0.733 & 0.867 & 0.900 & 0.900 & \textbf{1.000} & \textbf{1.000} & 0.945 & 0.933 & \textbf{1.000} \\
    irish & 500 & 6 & 2 & 1.1 & 66.7 & \underline{0.995} & 0.990 & 0.970 & \textbf{1.000} & \textbf{1.000} & 0.460 & 0.980 & 0.980 & 0.963 & 0.780 & \textbf{1.000} \\
    kc3 & 458 & 40 & 2 & 0 & 0 & \underline{0.935} & 0.912 & 0.912 & 0.901 & 0.890 & 0.902 & 0.859 & 0.902 & 0.913 & 0.891 & \textbf{0.961} \\
    kidney & 76 & 7 & 2 & 0 & 42.9 & \underline{0.920} & 0.667 & 0.533 & 0.600 & 0.733 & 0.750 & 0.875 & 0.688 & 0.877 & 0.688 & \textbf{0.968} \\
    LendingClub & 7370 & 53 & 12 & 3 & 22.6 & \underline{0.680} & 0.507 & 0.561 & 0.543 & 0.547 & 0.503 & 0.565 & 0.546 & 0.611 & 0.579 & \textbf{0.756} \\
    magic04 & 7450 & 11 & 2 & 0 & 9.1 & 0.810 & 0.805 & 0.823 & 0.831 & 0.821 & 0.787 & \underline{0.895} & 0.859 & 0.845 & 0.793 & \textbf{0.938} \\
    mammographic & 961 & 6 & 2 & 2.8 & 0 & \underline{0.865} & 0.823 & 0.812 & 0.812 & 0.818 & 0.736 & 0.824 & 0.829 & 0.827 & 0.829 & \textbf{0.913} \\
    move\_libras & 360 & 91 & 15 & 0 & 0 & \underline{0.905} & 0.569 & 0.639 & 0.653 & 0.694 & 0.583 & 0.819 & 0.806 & 0.887 & 0.750 & \textbf{0.928} \\
    no2 & 500 & 8 & 2 & 0 & 12.5 & \underline{0.775} & 0.590 & 0.440 & 0.610 & 0.620 & 0.610 & 0.710 & 0.740 & 0.634 & 0.500 & \textbf{0.819} \\
    phoneme & 5404 & 6 & 2 & 0 & 0 & 0.905 & 0.761 & 0.746 & 0.753 & 0.778 & 0.860 & 0.897 & \underline{0.911} & 0.865 & 0.736 & \textbf{0.927} \\
    plasma\_retinol & 315 & 14 & 2 & 0 & 28.6 & \underline{0.725} & 0.476 & 0.476 & 0.540 & 0.476 & 0.444 & 0.587 & 0.524 & 0.627 & 0.540 & \textbf{0.864} \\
    pm10 & 500 & 8 & 2 & 0 & 12.5 & \underline{0.790} & 0.550 & 0.550 & 0.550 & 0.560 & 0.500 & 0.630 & 0.660 & 0.627 & 0.590 & \textbf{0.928} \\
    schizo & 340 & 14 & 2 & 17.5 & 21.4 & \underline{0.745} & 0.574 & 0.544 & 0.632 & 0.618 & 0.485 & 0.588 & 0.588 & 0.641 & 0.677 & \textbf{0.804} \\
    Skin\_NonSkin & 7122 & 4 & 2 & 0 & 0 & 0.988 & 0.923 & 0.912 & 0.933 & 0.929 & 0.970 & \underline{0.997} & \textbf{0.999} & 0.969 & 0.922 & \textbf{0.999} \\
    socmob & 1156 & 6 & 2 & 0 & 83.3 & \underline{0.955} & 0.931 & 0.944 & 0.922 & 0.935 & 0.905 & 0.922 & 0.944 & 0.921 & 0.914 & \textbf{0.969} \\
    solar-flare & 1066 & 13 & 6 & 0 & 23.1 & 0.730 & 0.746 & 0.723 & 0.742 & 0.746 & 0.715 & 0.776 & 0.724 & \underline{0.799} & 0.738 & \textbf{0.902} \\
    broken\_machine & 7490 & 59 & 2 & 9.4 & 0 & 0.550 & 0.687 & 0.688 & 0.688 & 0.689 & 0.180 & 0.687 & \underline{0.690} & 0.649 & 0.688 & \textbf{0.831} \\
    triazines & 186 & 61 & 2 & 0 & 1.6 & 0.650 & 0.676 & 0.649 & 0.568 & 0.595 & 0.684 & 0.737 & 0.684 & \underline{0.754} & 0.711 & \textbf{0.903} \\
    veteran & 137 & 8 & 2 & 0 & 12.5 & 0.795 & 0.593 & 0.778 & 0.667 & 0.593 & 0.607 & \underline{0.821} & 0.750 & 0.757 & 0.607 & \textbf{0.845} \\
    weatherAUS & 7357 & 25 & 2 & 9.1 & 24 & 0.997 & 0.937 & 0.971 & 0.975 & 0.995 & 0.783 & \textbf{1.000} & \textbf{1.000} & 0.964 & 0.982 & \underline{0.998} \\
    wilt & 4839 & 6 & 2 & 0 & 16.7 & 0.975 & 0.944 & 0.967 & 0.973 & \underline{0.977} & 0.960 & 0.973 & \underline{0.977} & 0.953 & 0.962 & \textbf{0.991} \\
    Wine & 178 & 14 & 3 & 0 & 0 & \underline{0.989} & 0.971 & 0.971 & 0.971 & 0.971 & 0.972 & 0.916 & \textbf{1.000} & 0.949 & 0.972 & \textbf{1.000} \\
    \bottomrule
    \end{tabular}%
}
\label{tab:deepline}
\end{table*}
\begin{table*} 
\centering
\caption{Characteristics of the DiffPrep dataset collection and comparison of test accuracy.}
\vspace{1mm}

\setlength{\tabcolsep}{1.5pt}

\resizebox{\textwidth}{!}{%
    \begin{tabular}{l ccccc ccccccccccc}
    \toprule
    
     & \multicolumn{5}{c}{\textbf{Data Characteristics}} & \multicolumn{11}{c}{\textbf{Test Accuracy}} \\
    
    \cmidrule(lr){2-6} \cmidrule(lr){7-17}
    
    \textbf{Dataset} & \textbf{\#Inst.} & \textbf{\#Feat.} & \textbf{\#Lbl.} & \textbf{\%Miss.} & \textbf{\%Cat.} & \textbf{(ES$^*$)} & \textbf{DEF} & \textbf{RS} & \textbf{DP-Fix} & \textbf{DP-Flex} & \textbf{DL} & \textbf{HAI-AI} & \textbf{SAGA} & \textbf{CtxPipe} & \textbf{SwiftDP} & \textbf{Ours} \\
    \midrule
    
    abalone & 4177 & 9 & 28 & 0 & 12.50 & \underline{0.287} & 0.240 & 0.243 & 0.238 & 0.271 & 0.157 & 0.260 & 0.255 & \underline{0.287} & 0.274 & \textbf{0.522} \\
    ada\_prior & 4562 & 15 & 2 & 0.14 & 57.14 & \underline{0.857} & 0.848 & 0.844 & 0.853 & 0.846 & 0.803 & 0.801 & 0.833 & 0.818 & 0.834 & \textbf{0.920} \\
    avila & 20867 & 11 & 12 & 0 & 0 & \textbf{0.916} & 0.553 & 0.598 & 0.652 & 0.633 & 0.593 & 0.630 & 0.636 & 0.759 & 0.556 & \underline{0.837} \\
    connect-4 & 67557 & 43 & 3 & 0 & 0 & \underline{0.792} & 0.659 & 0.671 & 0.726 & 0.702 & 0.683 & 0.775 & 0.758 & 0.763 & 0.661 & \textbf{0.891} \\
    eeg & 14980 & 15 & 2 & 0 & 0 & \underline{0.861} & 0.589 & 0.658 & 0.675 & 0.683 & 0.607 & 0.556 & 0.658 & 0.740 & 0.575 & \textbf{0.885} \\
    google & 9367 & 9 & 2 & 2.19 & 50.00 & \underline{0.672} & 0.586 & 0.627 & 0.631 & 0.661 & 0.553 & 0.550 & 0.596 & 0.590 & 0.569 & \textbf{0.819} \\
    house & 1460 & 81 & 2 & 6.70 & 53.75 & \underline{0.955} & 0.928 & 0.938 & 0.932 & 0.952 & 0.771 & 0.928 & 0.913 & 0.818 & 0.938 & \textbf{0.960} \\
    jungle\_chess & 44819 & 7 & 3 & 0 & 0 & \underline{0.861} & 0.668 & 0.669 & 0.680 & 0.687 & 0.717 & 0.760 & 0.745 & \underline{0.861} & 0.676 & \textbf{0.872} \\
    micro & 20000 & 21 & 5 & 0 & 0 & \underline{0.634} & 0.564 & 0.579 & 0.595 & 0.593 & 0.613 & 0.633 & 0.556 & 0.605 & 0.573 & \textbf{0.793} \\
    mozilla4 & 15545 & 6 & 2 & 0 & 0 & \underline{0.940} & 0.855 & 0.922 & 0.924 & 0.927 & 0.747 & 0.870 & 0.932 & \underline{0.940} & 0.849 & \textbf{0.966} \\
    obesity & 2111 & 17 & 7 & 0 & 50.00 & \textbf{0.927} & 0.775 & 0.841 & 0.891 & 0.874 & 0.590 & 0.768 & 0.751 & 0.868 & 0.870 & \underline{0.911} \\
    page-blocks & 5473 & 11 & 5 & 0 & 0 & \underline{0.973} & 0.942 & 0.959 & 0.959 & \underline{0.973} & 0.940 & 0.935 & 0.849 & 0.965 & 0.964 & \textbf{0.981} \\
    pbcseq & 1945 & 19 & 2 & 4.13 & 22.22 & \underline{0.866} & 0.710 & 0.730 & 0.728 & 0.725 & 0.680 & 0.733 & \underline{0.866} & 0.805 & 0.679 & \textbf{0.874} \\
    pol & 15000 & 49 & 2 & 0 & 0 & \underline{0.977} & 0.884 & 0.879 & 0.903 & 0.916 & 0.873 & 0.916 & 0.888 & 0.949 & 0.887 & \textbf{0.990} \\
    run\_or\_walk & 88588 & 7 & 2 & 0 & 0 & \textbf{0.990} & 0.719 & 0.829 & 0.903 & 0.912 & 0.820 & 0.915 & 0.832 & 0.956 & 0.715 & \underline{0.976} \\
    shuttle & 58000 & 10 & 7 & 0 & 0 & \textbf{1.000} & 0.964 & 0.996 & 0.998 & \underline{0.999} & 0.790 & 0.951 & 0.405 & \textbf{1.000} & 0.965 & 0.998 \\
    uscensus & 32561 & 15 & 2 & 0.93 & 57.14 & \underline{0.854} & 0.848 & 0.840 & \underline{0.854} & 0.852 & 0.813 & 0.807 & 0.835 & 0.845 & 0.829 & \textbf{0.921} \\
    wall-robot-nav & 5456 & 25 & 4 & 0 & 0 & \underline{0.962} & 0.697 & 0.872 & 0.905 & 0.913 & 0.927 & 0.896 & 0.841 & 0.946 & 0.692 & \textbf{0.978} \\
    \bottomrule
    \end{tabular}%
}
\label{tab:diffprep}
\end{table*}





\end{document}